\documentclass[pdflatex,sn-mathphys-num]{sn-jnl}

\usepackage{graphicx}
\usepackage{multirow}
\usepackage{amsmath,amssymb,amsfonts}
\usepackage{amsthm}
\usepackage{mathrsfs}
\usepackage[title]{appendix}
\usepackage{xcolor}
\usepackage{textcomp}
\usepackage{manyfoot}
\usepackage{booktabs}
\usepackage{algorithm}
\usepackage{algorithmicx}
\usepackage{algpseudocode}
\usepackage{listings}
\usepackage{siunitx}
\usepackage{geometry}
\usepackage{bm} 

\theoremstyle{thmstyleone}

\theoremstyle{thmstyletwo}

\theoremstyle{thmstylethree}

\newcommand{\NB}[1]{\mathcal{N}_{\mathcal B}\!\left[#1\right]}

\begin{document}

\title{Phase-space entropy at acquisition reflects downstream learnability}

\author[2,3]{\fnm{Xiu-Cheng} \sur{Wang}}\email{xcwang\_1@stu.xidian.edu.cn}
\equalcont{Both authors contributed equally to this work.}

\author*[1]{\fnm{Jun-Jie} \sur{Zhang}}\email{zjacob@mail.ustc.edu.cn}
\equalcont{Both authors contributed equally to this work.}

\author[2,3]{\fnm{Nan} \sur{Cheng}}\email{dr.nan.cheng@ieee.org}

\author[4]{\fnm{Long-Gang} \sur{Pang}}\email{lgpang@ccnu.edu.cn}

\author[1]{\fnm{Taijiao} \sur{Du}}\email{dutaijiao@nint.ac.cn}

\author[5,6]{\fnm{Deyu} \sur{Meng}}\email{dymeng@mail.xjtu.edu.cn}

\affil[1]{\orgname{Northwest Institute of Nuclear Technology}, \orgaddress{\street{No. 28 Pingyu Road}, \city{Xi'an}, \postcode{710024}, \state{Shaanxi}, \country{China}}}

\affil[2]{\orgdiv{School of Telecommunications Engineering}, \orgname{Xidian University}, \orgaddress{\street{No. 2 South Taibai Road}, \city{Xi'an}, \postcode{710071}, \state{Shaanxi}, \country{China}}}
\affil[3]{\orgname{State Key Laboratory of ISN}, \orgaddress{\street{No. 2 South Taibai Road}, \city{Xi'an}, \postcode{710071}, \state{Shaanxi}, \country{China}}}

\affil[4]{\orgdiv{Key Laboratory of Quark and Lepton Physics (MOE) \& Institute of Particle Physics}, \orgname{Central China Normal University}, \orgaddress{\street{No. 152 Luoyu Road}, \city{Wuhan}, \postcode{430079}, \state{Hubei}, \country{China}}}

\affil[5]{\orgdiv{Ministry of Education Key Lab of Intelligent Networks and Network Security}, \orgname{Xi’an Jiaotong University}, \orgaddress{\street{No. 28 Xianning West Road}, \city{Xi'an}, \postcode{710049}, \state{Shaanxi}, \country{China}}}

\affil[6]{\orgdiv{School of Mathematics and Statistics}, \orgname{Xi’an Jiaotong University}, \orgaddress{\street{No. 28 Xianning West Road}, \city{Xi'an}, \postcode{710049}, \state{Shaanxi}, \country{China}}}


\abstract{
Modern learning systems work with data that vary widely across domains, but they all ultimately depend on how much structure is already present in the measurements before any model is trained. This raises a basic question: is there a general, modality-agnostic way to quantify how acquisition itself preserves or destroys the information that downstream learners could use? Here we propose an acquisition-level scalar $\Delta S_{\mathcal B}$ based on instrument-resolved phase space.
Unlike pixelwise distortion or purely spectral errors that often saturate under aggressive undersampling, $\Delta S_{\mathcal B}$ directly quantifies how acquisition mixes or removes joint space--frequency structure at the instrument scale. We show theoretically that \(\Delta S_{\mathcal B}\) correctly identifies the phase-space coherence of periodic sampling as the physical source of aliasing, recovering classical sampling-theorem consequences. Empirically, across masked image classification, accelerated MRI, and massive MIMO (including over-the-air measurements), $|\Delta S_{\mathcal B}|$ consistently ranks sampling geometries and predicts downstream reconstruction/recognition difficulty \emph{without training}. In particular, minimizing $|\Delta S_{\mathcal B}|$ enables zero-training selection of variable-density MRI mask parameters that matches designs tuned by conventional pre-reconstruction criteria.
These results suggest that phase-space entropy at acquisition reflects downstream learnability, enabling pre-training selection of candidate sampling policies and as a shared notion of information preservation across modalities.}


\maketitle

Data in modern machine learning are often treated as fixed objects to be optimized over once collected \cite{Goodfellow2016DeepLearning, LeCun2015DeepLearning, Ng2021DataCentricAI, Zha2023DataCentricAI}. Yet before any model is trained, those data are produced by physical instruments that sample continuous fields under hard constraints on time, bandwidth, and hardware complexity. This raises a basic acquisition-side question: \emph{can we quantify, in a modality-agnostic and pre-training way, whether a sampling/masking policy preserves the structure that downstream learning or reconstruction will rely on?}

Several mature lines of work already connect sensing choices to performance. Compressed sensing links recoverability to structural priors such as sparsity and to properties of the measurement operator \cite{Donoho2006CS, CandesWakin2008IntroCS, CandesRombergTao2006Stable, CandesTao2006RUP, Mallat2009Wavelet, RechtFazelParrilo2010}. Optimal experimental design optimizes measurement placement for a specified parametric model via information criteria \cite{Pukelsheim2006OptimumDesign, Atkinson2007OptimumDesign, Cover2006InfoTheory, KrauseGuestrin2005GP, Krause2008Submodular}. Data-centric AI and data valuation methods, in turn, quantify sample utility relative to a particular trained model and task \cite{KohLiang2017Influence, Ghorbani2019DataShapley, Mirzasoleiman2020CRAIG, Killamsetty2020GLISTER, Killamsetty2021GradMatch, Nguyen2020LEEP, You2021LogME}. Here, we focus on a complementary angle at acquisition time: can we compute a single, model-agnostic scalar directly from measurements (or a lightweight reference protocol) that summarizes how a sampling operator preserves the joint structure that downstream learners will later exploit?

Our approach uses phase space to describe what acquisition does to structure at the instrument’s resolution. Rather than measuring only \emph{how much} energy is kept, we quantify \emph{how} acquisition redistributes energy jointly over space and spatial frequency. Concretely, we construct an instrument-aligned, nonnegative phase-space density $\rho_I(\mathbf x;\mathbf k)$ using a Husimi/spectrogram representation (i.e., a smoothed Wigner distribution with a fixed resolution kernel) \cite{Wigner1932, Husimi1940, Cohen1989, Flandrin1999, Schleich2011QuantumOptics}. From $\rho_I$ we compute a band-normalized Shannon entropy $S_{\mathcal B}[I]$ over a Nyquist band $\mathcal B$, and we summarize the effect of an acquisition operator $\mathcal A$ by the band-entropy change
\[
\Delta S_{\mathcal B}
=
S_{\mathcal B}[I_{\mathrm{acq}}]
-
S_{\mathcal B}[I_{\mathrm{ref}}],
\qquad
I_{\mathrm{acq}}=\mathcal A[I].
\]
Here $I_{\mathrm{ref}}$ denotes a suitable reference (e.g., fully sampled or high-fidelity), and the same instrument kernel is used for both terms.
Intuitively, $S_{\mathcal B}$ is a coarse-grained, instrument-resolved notion of phase-space disorder: transformations that are effectively information-preserving at the measurement resolution should not produce a systematic drift of this entropy across a population, whereas masking, subsampling, truncation, or noise tend to induce mixing or removal in phase space \cite{Arnold1989ClassicalMech, 10.1093/nsr/nwae141, Zhang2025iScience}.

This perspective yields a simple phase-space acquisition principle: sampling strategies that produce smaller $|\Delta S_{\mathcal B}|$ tend to preserve the phase-space organization that downstream learners can exploit. Crucially, this principle is \emph{pre-training}: $\Delta S_{\mathcal B}$ is computed from the measurements themselves, without fitting a task model. We first make the mechanism explicit on the classical contrast between periodic (coherent) and random (incoherent) sampling. In spatial-domain masking, periodic lattices deterministically fold high-frequency content into the Nyquist band, increasing spectral mixing and thus band entropy, whereas random masks break lattice coherence and preserve band entropy in expectation up to vanishing fluctuations. This connects aliasing and incoherence to a single, instrument-resolved phase-space disorder measure, rather than to a task- or model-specific criterion.

We then test whether this acquisition-level scalar tracks downstream difficulty across heterogeneous modalities. In masked visual recognition \cite{He2022MAE, Bao2022BEiT, Pathak2016ContextEncoders, Dosovitskiy2021An, Xie2022SimMIM}, $\Delta S_{\mathcal B}$ separates periodic and random patch subsampling even in deep undersampling regimes where common distortion proxies become weakly informative. In accelerated \emph{magnetic resonance imaging} (MRI) \cite{Lustig2007SparseMRI, KakSlaney2001CT}, where undersampling occurs in $k$-space and entropy typically decreases due to missing coefficients, we use $|\Delta S_{\mathcal B}|$ as a \emph{design objective} to select variable-density mask parameters \emph{without training} and then verify the resulting downstream reconstruction gains. Finally, in \emph{massive multiple-input multiple-output} (MIMO) channel estimation \cite{Marzetta2010MassiveMIMO, Jose2011PilotContam, Larsson2014CommMag, Alkhateeb2014JSTSP, Gao2016JSAC}, we show that the same entropy auditing predicts reconstruction error in simulation and remains predictive in over-the-air measurements, indicating robustness to real hardware impairments.

\paragraph*{Contributions}
\begin{enumerate}
    \item \textbf{An acquisition-time phase-space entropy principle.}
    We identify and formalize the central phenomenon: for a broad class of sampling/masking operations, the difficulty of downstream reconstruction or recognition is  shaped \emph{at acquisition} and is reflected by how acquisition perturbs instrument-resolved phase-space organization. In this view, phase-space entropy at acquisition can reflect downstream learnability.

    \item \textbf{A phase-space framework that explains the mechanism and recovers classical consequences.}
    We provide an analysis toolkit based on Husimi-aligned, band-normalized phase-space entropy and use it to expose the physical mechanism behind structured aliasing: periodic subsampling induces a convex-mixture (spectral folding) effect in a Nyquist band, leading to an entropy increase via Jensen-type arguments, whereas randomized subsampling preserves entropy in expectation up to vanishing finite-size fluctuations. Classical sampling intuitions thus emerge as limiting consequences of the same phase-space entropy principle, rather than from a task-specific metric.

    \item \textbf{A practical, pre-training scalar with unique design value across modalities.}
    Building on this principle and theory, we derive a single actionable scalar $|\Delta S_{\mathcal B}|$ that can be computed \emph{before training} to rank sampling policies and predict downstream performance. Across domain experiments show a consistent monotone association with downstream learnability and provides discriminative power in regimes where conventional proxies saturate. Importantly, it enables \emph{zero-training} acquisition design---for example, selecting variable-density MRI mask parameters by minimizing $|\Delta S_{\mathcal B}|$ without training a reconstruction network.
\end{enumerate}

\section{Theoretical framework: phase-space entropy as an acquisition principle}

\subsection{Phase-space formulation and an entropy-based acquisition principle}

We provide the theoretical mechanism behind the \emph{phase-space entropy principle}, which yields direct, testable predictions for the downstream trends observed in vision, MRI and MIMO.

Let $I(\mathbf x)$ denote a real- or complex-valued field over spatial coordinates $\mathbf x$. The Wigner distribution $W_I(\mathbf x;\mathbf k)$ provides a joint representation in space and spatial frequency \cite{Wigner1932, Flandrin1999, Schleich2011QuantumOptics}, but it can be sign-indefinite and contains oscillations that are not directly observable at finite resolution. We therefore adopt an instrument-aligned, nonnegative Husimi (spectrogram-type) density \cite{Husimi1940, Cohen1989}:
\begin{equation}
\rho_I(\mathbf x;\mathbf k)
=
(\Phi * W_I)(\mathbf x;\mathbf k),
\label{eq:husimi-main}
\end{equation}
where $\Phi$ is a fixed nonnegative smoothing kernel encoding the joint space--frequency resolution of the measurement system, and $*$ denotes convolution over phase space. In numerical implementations, $\Phi$ is realized by a Gaussian-windowed short-time Fourier transform (STFT) at a resolution matched to the instrument; details are given in Methods.

To isolate spectral \emph{shape} (rather than overall energy), we normalize $\rho_I$ over a Nyquist band $\mathcal B$:
\begin{equation}
\NB{\rho_I}(\mathbf x;\mathbf k)
=
\frac{\rho_I(\mathbf x;\mathbf k)}{\displaystyle \int_{\mathcal B} \rho_I(\mathbf x;\tilde{\mathbf k})\,\mathrm d\tilde{\mathbf k}},
\quad
\mathbf k\in\mathcal B,
\label{eq:band-normalization}
\end{equation}
and define the local band entropy
\begin{equation}
s_{\mathcal B}(\mathbf x)
=
- \int_{\mathcal B}
\NB{\rho_I}(\mathbf x;\mathbf k)
\log \NB{\rho_I}(\mathbf x;\mathbf k)\,
\mathrm d\mathbf k,
\label{eq:local-entropy-main}
\end{equation}
with the global band entropy obtained by spatial averaging,
\begin{equation}
S_{\mathcal B}[I]
=
\int s_{\mathcal B}(\mathbf x)\,w(\mathbf x)\,\mathrm d\mathbf x,
\label{eq:global-entropy-main}
\end{equation}
where $w(\mathbf x)\ge 0$ is a weight. In practice, the spatial aggregation weight can be uniform or energy-weighted. Unless otherwise stated we use uniform weighting; for severely masked signals we use an energy-weighted (‘masked-safe’) aggregation to avoid instability in low-energy windows. Entropy here quantifies how mixed the accessible spectrum is at the instrument scale \cite{Shannon1948, Wehrl1978, Zurek1994}.

Acquisition is modeled as an operator $\mathcal A$ acting on the underlying field, producing $I_{\mathrm{acq}}=\mathcal A[I]$. Using the same $(\Phi,\mathcal B)$, we define the band-entropy change
\begin{equation}
\Delta S_{\mathcal B}
=
S_{\mathcal B}[I_{\mathrm{acq}}]
-
S_{\mathcal B}[I_{\mathrm{ref}}],
\label{eq:deltaS-main}
\end{equation}
where $I_{\mathrm{ref}}$ is a suitable reference (fully sampled or high-fidelity). These considerations motivate a (complete theory is provided in SI Appendix)

\begin{quote}
\noindent
\textbf{Phase-space-entropy-based acquisition principle.}
At a fixed instrument resolution, the magnitude of the acquisition-induced
band-entropy change $|\Delta S_{\mathcal B}|$ summarizes how strongly acquisition
disrupts joint space--frequency organization. \emph{Smaller $|\Delta S_{\mathcal B}|$
reflects easier downstream learnability}, whereas
larger $|\Delta S_{\mathcal B}|$ indicates greater acquisition-induced information loss.
\end{quote}

\subsection{Mechanism behind coherent folding versus incoherent perturbations}

We use the phase-space entropy framework to re-derive a widely accepted contrast---periodic (coherent) versus randomized (incoherent) sampling---as a transparent validation of the theory’s interpretability and portability. The goal here is to show that the instrument-resolved phase-space tool recovers the classical aliasing/coherence picture with a short chain of standard ingredients. 

\paragraph*{Masking acts as phase-space mixing at fixed instrument resolution}
Consider spatial-domain masking by pointwise multiplication: given a field $I(\mathbf x)$ and a (possibly complex-valued) mask $M(\mathbf x)$, the acquired field is $J(\mathbf x)=M(\mathbf x)\,I(\mathbf x)$. The Wigner distribution obeys a product--convolution law \cite{Wigner1932, Cohen1989, Flandrin1999}:
\begin{equation}
W_J(\mathbf x;\mathbf k)
=
\frac{1}{(2\pi)^2}
\int W_M(\mathbf x;\mathbf q)\,
W_I(\mathbf x;\mathbf k-\mathbf q)\,\mathrm d\mathbf q.
\label{eq:prod-conv-main}
\end{equation}
After applying the same Husimi smoothing kernel $\Phi$, and under a standard instrument-scale local-averaging approximation, the masking effect lifts to the Husimi level as
\begin{equation}
\rho_J(\mathbf x;\mathbf k)
\approx
\frac{1}{(2\pi)^2}
\bigl(K_M *_{\mathcal B} \rho_I\bigr)(\mathbf x;\mathbf k),
\label{eq:husimi-conv-main}
\end{equation}
where $K_M=\Phi*W_M$ is an effective nonnegative mixing kernel over $\mathbf k\in\mathcal B$. Thus, at fixed instrument resolution, acquisition modifies $I$ primarily by \emph{mixing} its pre-acquisition spectrum within the Nyquist band.

\paragraph*{Coherent periodic masks increase band entropy; incoherent random masks preserve it in expectation}
For periodic subsampling, the effective kernel $K_M$ concentrates near reciprocal-lattice frequencies. After instrument-scale averaging, the resulting band-normalized spectrum within a Nyquist band becomes a convex mixture of shifted copies of the original spectrum,
\begin{equation}
\NB{\rho_J}(\mathbf x;\mathbf k)
\approx
\sum_{\boldsymbol\kappa_{\mathbf n}\in\mathcal B} w_{\mathbf n}\,
\NB{\rho_I}\bigl(\mathbf x;\mathbf k-\boldsymbol\kappa_{\mathbf n}\bigr),
\label{eq:convex-mixture-main}
\end{equation}
where $\sum_{\mathbf n} w_{\mathbf n}=1,\ \ w_{\mathbf n}\ge 0,$ and Jensen’s inequality implies
\begin{equation}
s_{\mathcal B}^{(J)}(\mathbf x)\ge s_{\mathcal B}^{(I)}(\mathbf x)
\quad\Rightarrow\quad
S_{\mathcal B}[J]\ge S_{\mathcal B}[I],
\end{equation}
with strict inequality except in degenerate ``lattice-invariant'' cases (details are provided in SI Appendix). This recasts classical aliasing in phase-space terms: coherent sampling creates structured spectral folding inside $\mathcal B$, increasing spectral mixing and therefore increasing instrument-resolved band entropy.

Randomized subsampling breaks lattice coherence. A convenient analytical model is an i.i.d.\ Bernoulli mask\footnote{The theoretical analysis uses Bernoulli sampling because it yields closed-form expectations, whereas the experiments use periodic sampling without replacement under fixed budgets. For moderate budgets these two behave similarly in entropy.}. After smoothing and averaging over many instrument-scale cells, the induced perturbation does not create systematic folding into preferred low-frequency regions. Quantitatively, the Husimi density is preserved in expectation and concentrates around its mean:
\begin{equation}
\mathbb E[\rho_J]
=
\rho_I + O\!\left(\frac{1}{N}\right),\ 
\mathbb E\|\rho_J - \rho_I\|_{L^1(\mathcal B)}
=
O\!\left(\frac{1}{\sqrt{N}}\right),
\end{equation}
where $N$ is the number of effectively independent instrument-scale cells. The $L^1$ concentration follows from standard bounds for sums of independent bounded contributions \cite{Tropp2015RandMatrices}; under a mild lower bound on local band energy, the band-normalization in \eqref{eq:band-normalization} is Lipschitz-stable, so the induced entropy perturbation inherits the same vanishing-fluctuation scaling. In short, the same phase-space tool distinguishes coherent folding (entropy increase) from incoherent perturbations (entropy preserved in expectation) using broadly standard ingredients.

\emph{Prediction for Experiments 1 and 3 (vision and MIMO).}
Under spatial-domain masking/subsampling at matched budgets, periodic geometries should yield larger $\Delta S_{\mathcal B}>0$ than randomized geometries, and configurations with smaller $\Delta S_{\mathcal B}$ should be easier for downstream learners.

\paragraph*{Remark: frequency-domain undersampling (MRI) and the role of $|\Delta S_{\mathcal B}|$}
The above mechanism explains the sign of $\Delta S_{\mathcal B}$ when acquisition acts primarily by coherent folding in the spatial domain. In frequency-domain undersampling (as in MRI), acquisition removes $k$-space coefficients rather than folding them, so $\Delta S_{\mathcal B}$ typically becomes negative. We therefore apply the same acquisition principle through the magnitude $|\Delta S_{\mathcal B}|$: among candidate $k$-space masks at a fixed acceleration factor, designs with smaller $|\Delta S_{\mathcal B}|$ (entropy closer to the reference) are predicted to better preserve instrument-resolved phase-space structure and to yield easier reconstructions.

\emph{Prediction for Experiment 2 (MRI).}
Among candidate $k$-space sampling designs, minimizing $|\Delta S_{\mathcal B}|$ should improve downstream reconstruction quality, enabling zero-training selection of variable-density mask parameters.

\section{Experimental verification: cross-modality validation in vision, MRI, and MIMO (with over-the-air measurements)}

We now evaluate the phase-space-entropy-based acquisition principle across three sensing modalities, ordered by increasing domain shift and implementation complexity. We start from masked image recognition, a visually intuitive setting where acquisition acts in the spatial domain and where pre-training distortion proxies are widely used. We then move to accelerated MRI, a canonical acquisition-design problem, to test not only predictive alignment with classical mask hierarchies but also whether $|\Delta S_{\mathcal B}|$ can \emph{close the loop} as a zero-training design objective before any network is trained. Finally, we consider massive MIMO wireless channel estimation and verify that the same acquisition-level entropy auditing remains predictive under realistic hardware impairments via over-the-air (OTA) measurements. Across all experiments, the emphasis is on what can be decided \emph{at acquisition}---before choosing a downstream model or paying the cost of training.

\subsection{Experiment 1: Vision --- phase-space entropy and masked image classification}

We first test the framework in a visual recognition setting where masking is performed directly in the spatial domain. The task is $100$-way classification on mini-ImageNet \cite{Vinyals2016Matching} using a Vision Transformer (ViT) \cite{Dosovitskiy2021An} under different patch-subsampling schemes. Images are resized to $256\times256$ pixels and decomposed into $16\times16$ non-overlapping patches. For each observation budget, defined by a sampling interval \(k\in\{2,4,8\}\) (keeping \(1/k^2\) of patches), we compare two geometries with identical budgets: a \emph{periodic} mask that retains patches on a regular lattice, and a \emph{random} mask that selects the same number of patches uniformly without replacement.

Our goal is to test a more specific claim: \emph{as acquisition becomes severely undersampled, conventional pre-training proxies become weakly discriminative, whereas $\Delta S_{\mathcal B}$ remains sensitive to geometry-induced structure disruption}. We compute \(\Delta S_{\mathcal B}\) using the multi-scale, energy-weighted protocol described in Methods, and benchmark it against two common pre-training distortion measures: spatial Peak Signal-to-Noise Ratio (PSNR) and Fourier-magnitude \(L_2\) distance.

Downstream classification accuracy confirms the predicted learnability hierarchy. As the subsampling interval increases from \(k=2\) to \(8\), test accuracy expectedly degrades, but Random sampling consistently outperforms Periodic sampling across all budgets. Specifically, Random sampling yields higher top-1 accuracy than Periodic sampling at \(k=2\) (\(30.3\%\) vs.\ \(26.2\%\)), \(k=4\) (\(18.9\%\) vs.\ \(13.5\%\)), and \(k=8\) (\(10.5\%\) vs.\ \(9.7\%\)). With this performance gap established, we examine whether acquisition-level metrics can anticipate it.

\begin{figure}[t!]
    \centering
    \includegraphics[width=\linewidth]{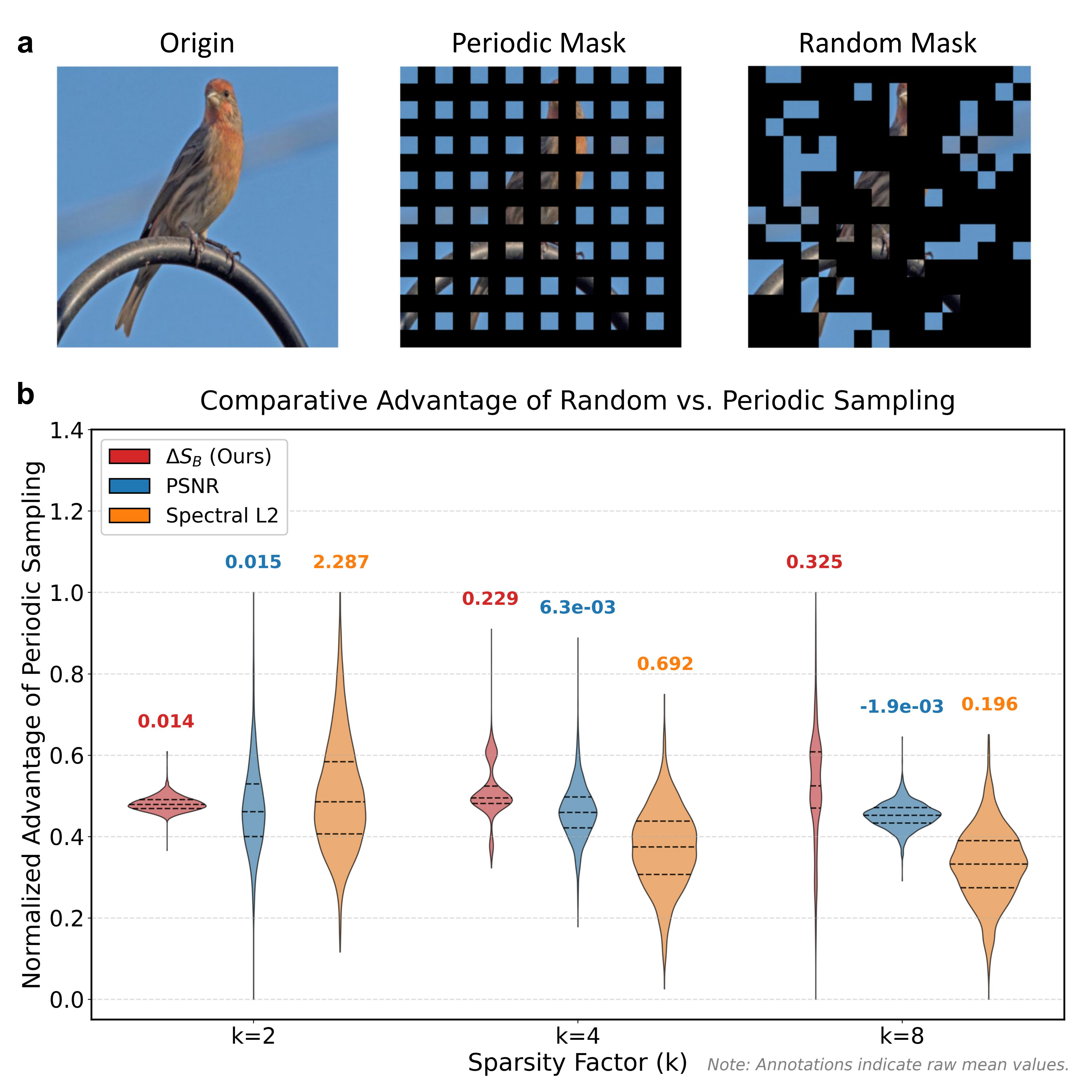}
    \caption{\textbf{Phase-space entropy uniquely discriminates the advantage of random over periodic sampling.}
    \textbf{(a)} Visual illustration of the sampling geometries, showing the original fully sampled signal (left), a periodic sampling mask (center), and a random sampling mask (right).
    \textbf{(b)} Violin plots quantifying the \emph{comparative advantage} provided by Random sampling over Periodic sampling across the test set for three metrics: \(\Delta S_{\mathcal B}\) (red), PSNR (blue), and Spectral \(L_2\) (orange). The y-axis represents the normalized advantage (e.g., \(\Delta S_{\mathcal B}^{\text{periodic}} - \Delta S_{\mathcal B}^{\text{random}}\)); higher positive values indicate a stronger preference for the Random geometry. Inner dashed lines mark the quartiles. While PSNR and Spectral \(L_2\) advantages concentrate near zero and show weak sensitivity to sparsity changes, the \(\Delta S_{\mathcal B}\) advantage is strictly positive and becomes increasingly pronounced as sparsity rises (\(k=2 \to 8\)). This confirms that phase-space entropy effectively captures the geometry-induced learnability gap that conventional pre-training proxies tend to compress.}
    \label{fig:vision}
\end{figure}

Figure~\ref{fig:vision}b summarizes the results by plotting the sample-wise \emph{comparative advantage} of random sampling over periodic sampling. For the entropy metric, we track the reduction in acquisition-induced perturbation,
\(\Delta S_{\mathcal B}^{\text{periodic}} - \Delta S_{\mathcal B}^{\text{random}}\),
so positive values indicate better phase-space structure preservation by random masks.

\begin{figure*}[t!]
    \centering
    \includegraphics[width=0.9\linewidth]{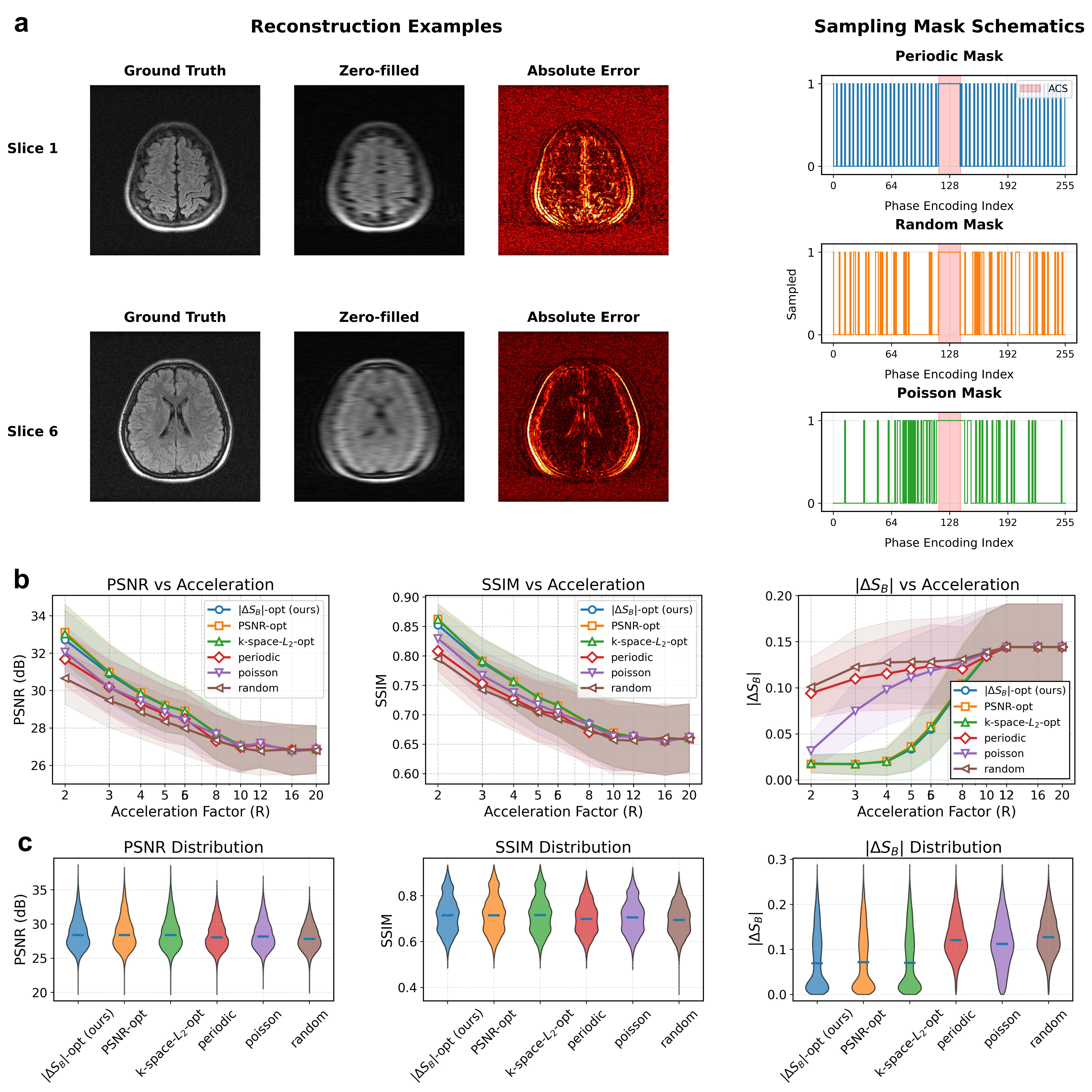}
    \caption{\textbf{Band-entropy change under $k$-space subsampling predicts reconstruction quality in accelerated MRI.}
    \textbf{(a)} Schematic of accelerated acquisition: binary masks along the phase-encoding ($k_y$) direction remove Fourier lines outside a central auto-calibration region. Zero-filled inverse FFT produces aliased magnitude images that serve as inputs to a U-Net.
    \textbf{(b)} Reconstruction PSNR/SSIM and average $|\Delta S_{\mathcal B}|$ for six mask families across acceleration factors. Canonical masks (Periodic, Random, Poisson) recover the familiar ranking, with Poisson achieving both the best reconstruction quality and the smallest $|\Delta S_{\mathcal B}|$. Three variable-density Cartesian designs are selected in a pre-training way by optimizing, respectively, $|\Delta S_{\mathcal B}|$, $k$-space $L_2$ error, or pre-reconstruction PSNR; all three attain strong PSNR/SSIM, and the $|\Delta S_{\mathcal B}|$-optimized mask consistently yields the smallest entropy perturbation.
    \textbf{(c)} Violin plots of test-set PSNR, SSIM, and $|\Delta S_{\mathcal B}|$ for all six masks. Reconstruction metrics are tightly clustered and show limited separation between acquisition geometries, whereas $|\Delta S_{\mathcal B}|$ exhibits pronounced differences in both mean and distribution shape, indicating that phase-space entropy reveals latent structure in acquisition-induced information loss beyond what standard image-domain metrics capture.}
    \label{fig:mri}
\end{figure*}

Three observations follow. First, the advantage measured by \(\Delta S_{\mathcal B}\) is consistently positive across budgets, aligning with the theoretical mechanism that periodic lattices induce coherent spectral folding while random masks do not. Second, the mean advantage increases substantially as the observation budget tightens (\(k=2\to 8\)), reflecting that coherent aliasing becomes increasingly dominant in the hard undersampling regime. Third, and most importantly for the acquisition principle, PSNR and Spectral \(L_2\) exhibit comparatively compressed advantage distributions at matched budgets, whereas \(\Delta S_{\mathcal B}\) retains a wider dynamic range. In other words, \(\Delta S_{\mathcal B}\) provides a pre-training signal with higher discriminative resolution for geometry-induced learnability than standard distortion proxies in this regime.

\subsection{Experiment 2: MRI --- zero-training acquisition design via phase-space entropy}

We next test the framework in accelerated MRI, where undersampling occurs in the \emph{frequency} domain ($k$-space). In this regime, removing Fourier lines reduces accessible spectral support and typically decreases entropy, so generally \(\Delta S_{\mathcal B} < 0\). We therefore use the \emph{magnitude} $|\Delta S_{\mathcal B}|$ as an acquisition-level proxy: sampling schemes that keep $|\Delta S_{\mathcal B}|$ closer to the fully sampled reference are predicted to be easier for downstream reconstruction.

We work with multi-coil complex-valued $k$-space data. Accelerated acquisition is emulated by applying 1D binary masks along the phase-encoding direction ($k_y$) outside a central auto-calibration region. A 2D U-Net \cite{Ronneberger2015UNet} reconstructs magnitude images from zero-filled inputs (Fig.~\ref{fig:mri}a). The central question here is not only whether the entropy metric aligns with known mask hierarchies, but whether it can \emph{guide acquisition design without training}: we select sampling parameters using only pre-training criteria, then train reconstruction networks \emph{afterward} to evaluate the predictive value of those choices. Finally, we compute \(|\Delta S_{\mathcal B}|\) between the fully sampled reference image and the zero-filled image obtained by applying the k-space mask and inverse FFT (before any learning).

We compare two categories of masks:
\begin{enumerate}
    \item \emph{Canonical masks:} standard periodic, random, and Poisson-disc--like variable-density patterns \cite{Lustig2007SparseMRI}.
    \item \emph{Parametric variable-density masks:} a family parameterized by \(p(k)\propto (1+\alpha |k-k_{\mathrm{center}}|)^{-\beta}\). We select \((\alpha,\beta)\) by optimizing three distinct pre-training criteria on calibration data: (i) minimizing \(|\Delta S_{\mathcal B}|\), (ii) minimizing $k$-space $L_2$ error, and (iii) maximizing zero-filled PSNR.
\end{enumerate}
Each selected design is then treated as a fixed acquisition geometry for training a dedicated reconstruction network.

Figure~\ref{fig:mri}b--c highlights two layers of validation. First, among canonical masks, Poisson sampling achieves the best reconstruction quality and the smallest $|\Delta S_{\mathcal B}|$, while periodic and random sampling induce larger entropy perturbations and lower performance. This confirms that the entropy metric is consistent with the established advantage of incoherent, variable-density sampling in accelerated MRI \cite{Lustig2007SparseMRI, KakSlaney2001CT}. Second, and crucially for acquisition design, the variable-density mask selected solely by \textbf{minimizing $|\Delta S_{\mathcal B}|$} (without any knowledge of the reconstruction network) yields reconstruction PSNR/SSIM that matches designs selected by $k$-space $L_2$ error or zero-filled PSNR. This establishes a pre-training design loop: entropy is computed \emph{before training}, used to select acquisition parameters, and the downstream reconstruction results validate the predicted ranking.

Beyond mean performance, Fig.~\ref{fig:mri}c shows that while PSNR and SSIM distributions of strong designs can be tightly clustered, $|\Delta S_{\mathcal B}|$ remains more structurally separated across mask families. This is the distinctive value in an acquisition context: $|\Delta S_{\mathcal B}|$ provides a physically grounded, model-agnostic scale for comparing acquisition policies even when conventional outcome metrics partially saturate and offer limited guidance for selection.

\subsection{Experiment 3: Wireless MIMO --- entropy auditing and over-the-air validation}

Finally, we test transfer to a markedly different domain: massive MIMO wireless channel estimation, where subsampling is performed over antenna elements and pilot resources rather than pixels or $k$-space. Accurate channel state information (CSI) is essential for multi-antenna communication performance, yet pilot overhead grows rapidly with array size. In practice, only a subset of channel coefficients can be observed, requiring reconstruction of the full channel matrix.

We consider narrowband MIMO channels represented by complex matrices $\mathbf H\in\mathbb C^{N_r\times N_t}$ \cite{Marzetta2010MassiveMIMO, Jose2011PilotContam, Larsson2014CommMag, Alkhateeb2014JSTSP, Gao2016JSAC}. Under a fixed pilot budget, only entries indexed by $\Omega\subseteq\{1,\ldots,N_r\}\times\{1,\ldots,N_t\}$ are observed, and a neural network estimates $\widehat{\mathbf H}$ from the subsampled observations. We use a periodic deactivation pattern with interval $d$, corresponding to a regular-lattice geometry in the antenna/pilot index space. We compare it to random subsampling under matched budgets.

We compute the Husimi-based band-entropy $S_{\mathcal B}$ on the magnitude field $|\mathbf H|$ using small 2D Gaussian windows spanning a few adjacent antennas. Simulated channels are generated using a clustered geometric model with realistic angular spreads and path gains, with additive noise at a fixed SNR. For each deactivation pattern and budget, we train a reconstruction network with the same architecture and loss; performance is measured by normalized mean squared error (NMSE). We then validate robustness via OTA measurements using a $1\times 8$ uniform linear array (ULA) at both transmitter and receiver, forming an $8\times 8$ MIMO link.

\begin{figure}[t!]
    \centering
    \includegraphics[width=\linewidth]{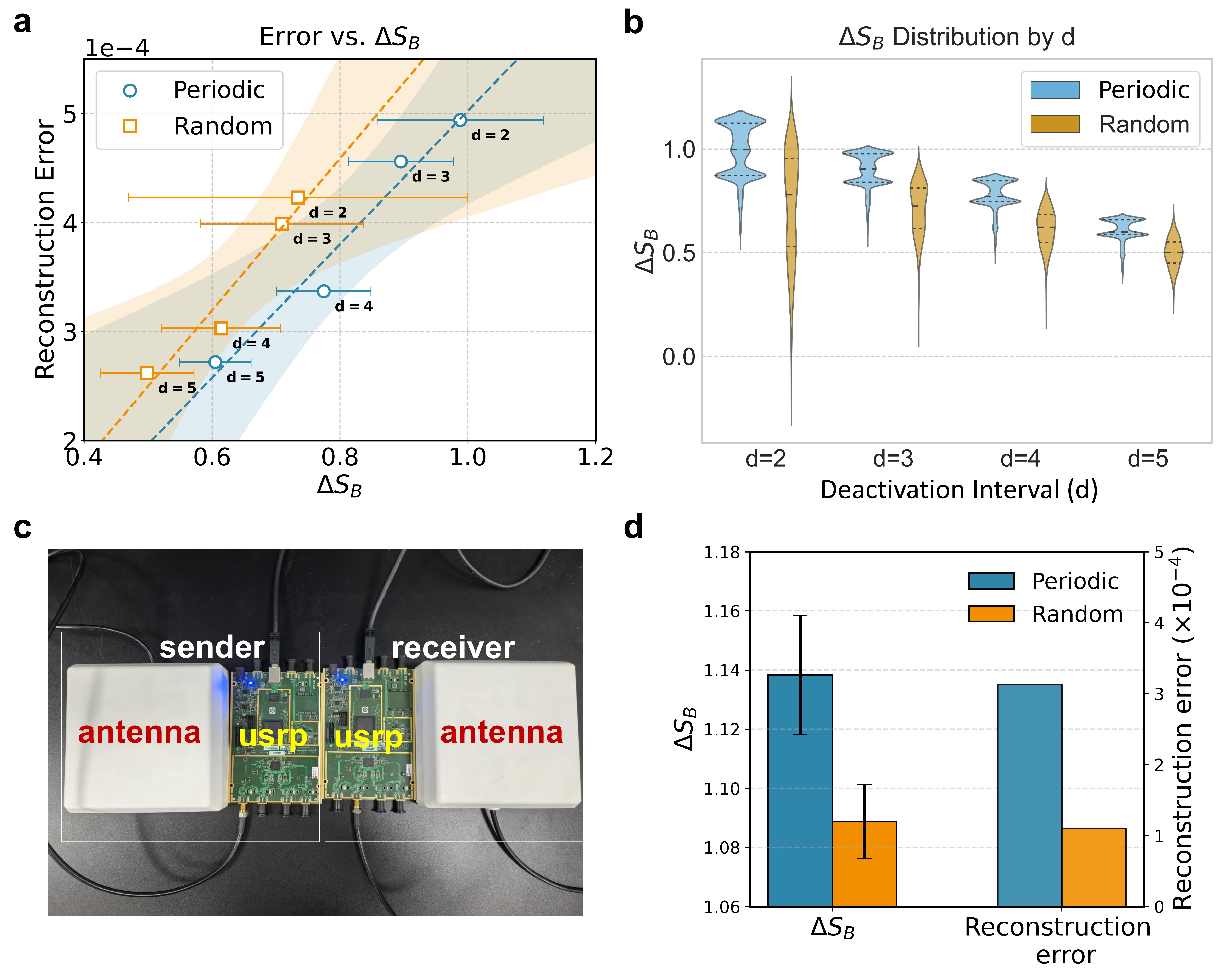}
    \caption{\textbf{Entropy auditing of MIMO channels and over-the-air validation.}
    \textbf{(a)} NMSE of CSI reconstruction versus mean band-entropy change $\Delta S_{\mathcal B}$ for Periodic and random antenna deactivation at different pilot budgets in simulation. Each point corresponds to a fixed deactivation interval $d$ and geometry; larger $\Delta S_{\mathcal B}$ is associated with higher NMSE.
    \textbf{(b)} Violin plots of $\Delta S_{\mathcal B}$ across simulated channels for each deactivation scheme and budget. Periodic deactivation consistently yields larger entropy perturbations than random deactivation at matched budgets.
    \textbf{(c)} Schematic of the OTA testbed with a $1\times 8$ ULA at the transmitter and a $1\times 8$ ULA at the receiver, forming an $8\times 8$ MIMO link.
    \textbf{(d)} OTA results: random deactivation achieves lower NMSE and smaller $|\Delta S_{\mathcal B}|$ than periodic deactivation under the same pilot budget.}
    \label{fig:mimo}
\end{figure}

As shown in Fig.~\ref{fig:mimo}a--b, periodic deactivation induces a larger positive $\Delta S_{\mathcal B}$ than random deactivation at matched budgets, consistent with stronger coherent structure disruption from regular-lattice removal patterns. Across budgets, NMSE increases with $\Delta S_{\mathcal B}$: configurations with larger entropy perturbations tend to be harder to reconstruct. Crucially, this association persists in OTA measurements (Fig.~\ref{fig:mimo}d), indicating that the acquisition-level entropy audit remains informative under practical hardware impairments. This final experiment therefore supports the intended use case of the principle: a single pre-training scalar computed from acquired measurements can anticipate downstream difficulty even when moving across domains and from simulation to real systems.

\section{Discussion}

We show that acquisition reshapes joint space--frequency organization in a way that can be audited at a fixed instrument resolution, and that the resulting band-entropy change $|\Delta S_{\mathcal B}|$ tracks downstream difficulty across vision, accelerated MRI, and massive MIMO (including OTA measurements). This places the proposed metric at a complementary layer to existing paradigms---compressed sensing, optimal design, and model-based data valuation---by explicitly targeting a pre-training, modality-agnostic diagnostic of how acquisition preserves or disrupts structure that learners later exploit.

\subsection{A unified view of sign and mechanism across modalities}
An important clarification is that $\Delta S_{\mathcal B}$ has a modality-dependent sign mechanism, while $|\Delta S_{\mathcal B}|$ provides a modality-agnostic ``distance-to-reference'' scale. In spatial-domain masking (vision/MIMO), coherent periodic geometries induce spectral folding within the Nyquist band and typically increase entropy ($\Delta S_{\mathcal B}>0$); in frequency-domain undersampling (MRI), acquisition removes coefficients and typically decreases entropy ($\Delta S_{\mathcal B}<0$). This links the sign of $\Delta S_{\mathcal B}$ to a simple mechanism (folding vs. removal).

\subsection{Relation to existing theories and design criteria}
The phase-space entropy view complements classical compressed sensing and incoherence arguments by making ``coherence'' visible at the level of instrument-resolved phase-space mixing. Compared with outcome-driven pre-reconstruction proxies (e.g., PSNR of zero-filled images) or purely spectral discrepancies that may saturate in severe undersampling, $|\Delta S_{\mathcal B}|$ retains discriminative power because it measures how acquisition redistributes energy jointly in space and frequency, not only how much energy is retained, making it a practical acquisition-time interface with minimal assumptions about the learner.

\subsection{Practical usage: a pre-training acquisition-selection recipe}
A key advantage of $|\Delta S_{\mathcal B}|$ is that it enables ``design before training.'' In practice, one may use the following workflow:
\begin{enumerate}
\item \textbf{Define a lightweight reference protocol} $I_{\mathrm{ref}}$: fully sampled data when available, or a high-fidelity subset (e.g., a short calibration scan / ACS-like region) collected once per instrument setting.
\item \textbf{Choose an instrument-matched analysis scale} $(\texttt{win},\sigma,\texttt{hop})$: windows should resolve the scale at which acquisition artifacts manifest and at which downstream models extract features. When scale is uncertain (e.g., natural images under heavy masking), multi-scale aggregation can stabilize rankings.
\item \textbf{Enumerate candidate acquisition policies} (masks, sampling geometries, pilot/antenna patterns), compute $|\Delta S_{\mathcal B}|$ on a calibration set, and \textbf{select a small set of top-ranked policies} for downstream training or deployment.
\end{enumerate}
This yields a practical pre-training loop for screening acquisition policies.

\subsection{Limitations and failure modes}
$|\Delta S_{\mathcal B}|$ is an acquisition-level ordering statistic rather than a universal performance guarantee; it is most appropriate for ranking candidate policies under a consistent reference protocol and analysis scale. First, $|\Delta S_{\mathcal B}|$ is defined relative to a reference; if the reference is noisy, biased, or collected under mismatched instrument conditions, entropy differences can reflect reference mismatch rather than acquisition quality. Use a consistent reference protocol and treat $|\Delta S_{\mathcal B}|$ as a population-level statistic. Second, the analysis depends on an instrument-resolved scale: overly small windows may under-resolve coherent artifacts (e.g., periodic folding patterns), while overly large windows may average away localized structure. When the appropriate instrument scale is uncertain, we recommend a staircase multi-scale audit (as in Vision Experiment): compute rankings across a ladder of window sizes and select either (i) the smallest scale within a stability plateau (high rank correlation across neighboring scales), or (ii) an energy-weighted multi-scale aggregate. Moreover, our ablations suggest that qualitative rankings are stable across broad parameter ranges when the analysis scale is physically and architecturally consistent, but extreme mismatches can lead to reversals. Third, $|\Delta S_{\mathcal B}|$ quantifies generic structure disruption rather than task-specific sufficiency: tasks that depend disproportionately on specific bands or features may require band- or region-weighted variants of the metric.

\subsection{Outlook}
Beyond the three modalities studied here, an instrument-resolved phase-space entropy audit offers a compact way to compare acquisition quality across heterogeneous sensors using a consistent reference protocol. This may enable principled allocation of sensing resources (time, bandwidth, pilot overhead, sampling density) and pre-training screening of candidate policies in multi-sensor systems. An important direction is task-adapted variants (e.g., band- or region-weighted entropies) when downstream objectives emphasize specific structures. Another direction is closed-loop sensing, where $|\Delta S_{\mathcal B}|$ is computed on-the-fly to monitor and maintain acquisition quality under shifting conditions.

%

\section{Methods}

\subsection{Phase-space representation and band-entropy change}

In the main text we formulate acquisition in terms of a Husimi-type phase-space density that encodes how energy is jointly distributed over position $\mathbf x$ and spatial frequency $\mathbf k$ via
Eqs.~(\ref{eq:husimi-main})–(\ref{eq:global-entropy-main}). Detailed derivations, including the connection to the Wigner distribution and Liouville dynamics, are provided in the SI Appendix; here we summarize only the operational procedure used in our experiments.

Given a real- or complex-valued field $I(\mathbf x)$ sampled on a regular grid
(e.g., an image, a channel matrix, or an MRI magnitude image), we first form
local, windowed patches centered at spatial positions $\{\mathbf x_i\}$ with a
fixed window size \texttt{win} and stride \texttt{hop}. Each patch is multiplied
by a Gaussian window
\begin{equation}
g_\sigma(\mathbf u)
\;\propto\;
\exp\!\Bigl(-\tfrac{\|\mathbf u\|^2}{2\sigma^2}\Bigr),
\end{equation}
with standard deviation $\sigma$ in pixel or antenna-index units, and then
transformed by a discrete Fourier transform (DFT) with orthonormal
normalization. This yields a local power spectrum
\begin{equation}
P_I(\mathbf x_i;\mathbf k)
=
\bigl|\mathcal F\{g_\sigma(\cdot-\mathbf x_i)\,I(\cdot)\}(\mathbf k)\bigr|^2.
\end{equation}
Up to the choice of normalization, $P_I$ is a discrete spectrogram
approximating the Husimi density $\rho_I(\mathbf x;\mathbf k)$ in
Eq.~(\ref{eq:husimi-main}), with $(\texttt{win},\sigma,\texttt{hop})$
implementing the smoothing kernel $\Phi$ and the instrument’s effective
space–frequency resolution.

To obtain a band-limited phase-space probability density, we restrict the
discrete frequencies to the Nyquist band $\mathcal B$ and normalize $P_I$
over $\mathbf k\in\mathcal B$ at each spatial location:
\begin{equation}
\NB{\rho_I}(\mathbf x_i;\mathbf k)
=
\frac{P_I(\mathbf x_i;\mathbf k)}{\displaystyle
\sum_{\mathbf k'\in\mathcal B}P_I(\mathbf x_i;\mathbf k')},
\qquad
\mathbf k\in\mathcal B,
\end{equation}
which is the discrete analogue of Eq.~(\ref{eq:band-normalization}).
The corresponding local band-entropy is the Shannon entropy of this
distribution,
\begin{equation}
s_{\mathcal B}(\mathbf x_i)
=
-
\sum_{\mathbf k\in\mathcal B}
\NB{\rho_I}(\mathbf x_i;\mathbf k)\,
\log \NB{\rho_I}(\mathbf x_i;\mathbf k),
\end{equation}
approximating Eq.~(\ref{eq:local-entropy-main}). The global band-entropy
$S_{\mathcal B}[I]$ is then obtained by averaging over all window centers,
\begin{equation}
S_{\mathcal B}[I]
=
\frac{1}{N_{\mathrm{win}}}
\sum_{i=1}^{N_{\mathrm{win}}}
s_{\mathcal B}(\mathbf x_i),
\end{equation}
corresponding to Eq.~(\ref{eq:global-entropy-main}) with a uniform weight
$w(\mathbf x)$.

Given a reference field $I_{\mathrm{ref}}$ and an acquired field
$\,I_{\mathrm{acq}}=\mathcal A[I_{\mathrm{ref}}]\,$ produced by a sampling
operator $\mathcal A$ (e.g., spatial masking, $k$-space undersampling, or
antenna deactivation), we compute their respective band-entropies using the
\emph{same} window parameters $(\texttt{win},\sigma,\texttt{hop})$ and
frequency band $\mathcal B$, and define the band-entropy change
\begin{equation}
\Delta S_{\mathcal B}
=
S_{\mathcal B}[I_{\mathrm{acq}}]
-
S_{\mathcal B}[I_{\mathrm{ref}}],
\end{equation}
as in Eq.~(\ref{eq:deltaS-main}). In the vision and MIMO experiments,
$I_{\mathrm{ref}}$ and $I_{\mathrm{acq}}$ are taken directly in the spatial
domain (masked or unmasked images, fully sampled or subsampled channel
matrices). In the MRI experiments, $I_{\mathrm{ref}}$ and $I_{\mathrm{acq}}$
are formed by applying a centered, orthonormal inverse DFT to the fully
sampled and masked $k$-space data, respectively, followed by root-sum-of-squares
coil combination to obtain magnitude images.

\subsection{Vision: masked mini-ImageNet classification with a Vision Transformer}

\paragraph*{Dataset and preprocessing}
We use the mini-ImageNet dataset \cite{Vinyals2016Matching}, consisting of 100 classes with 600 RGB images per class (60{,}000 images total). Images are resized to \(256\times256\) pixels and normalized channel-wise using ImageNet statistics. We adopt the standard train/validation/test split used in prior work on few-shot and representation learning.

\paragraph*{Architecture and training}
We use a Vision Transformer (ViT-B/16) \cite{Dosovitskiy2021An} that processes each \(256\times256\) image as a sequence of non-overlapping \(16\times16\) patches (256 patches per image). The patch embeddings, class token, and positional encodings follow the standard ViT-B/16 configuration. For each acquisition condition (mask type and budget), a separate classifier head is trained from scratch, sharing the same ViT backbone architecture and hyperparameters.

We optimize with AdamW (learning rate \(10^{-3}\), weight decay \(0.05\)), a cosine learning-rate schedule with linear warmup, batch size 256, and a total of 300 epochs. Data augmentation includes random cropping and horizontal flipping; no CutMix or Mixup is used to avoid confounding the masking operation. Training and evaluation are repeated with three random seeds per condition; we report mean test accuracy and 95\% confidence intervals.

\paragraph*{Masking strategies}
To emulate subsampled acquisition, we apply patch-level masks to each image before feeding it to the ViT. For a given sampling interval \(k\in\{2,4,8\}\), we retain approximately \(1/k^2\) of the patches and mask out the rest. Two geometries are considered:
\begin{itemize}
    \item \textbf{Periodic mask:} patches are retained on a regular \(k\times k\) lattice over the \(16\times16\) patch grid, corresponding to periodic subsampling in both spatial directions.
    \item \textbf{Random mask:} the same number of patches is selected uniformly at random without replacement from the patch grid.
\end{itemize}
Masked patches are replaced by a learned mask token; the visible patches and mask tokens are then processed by the ViT in the usual way. For each \(k\), a single random mask is drawn and fixed for all training and evaluation runs to isolate the effect of geometry rather than mask-to-mask variability.

\paragraph*{Phase-space entropy computation for vision}
For entropy auditing, we form the acquired field by zero-filling missing patches (i.e., masked patch pixels are set to 0); this is separate from the learned mask token used as model input. To ensure the entropy metric \(\Delta S_{\mathcal B}\) is robust across varying sparsity levels (particularly for large \(k\) where many local windows may be empty), we implement a \textbf{multi-scale, energy-weighted} Husimi analysis. Instead of relying on a single window size, we compute the band-entropy across a spectrum of scales defined by window sizes \(W \in \{32, 48, \dots, 256\}\) pixels, with corresponding Gaussian widths \(\sigma \approx W/6\). 

For each scale, the local band-entropy \(s_{\mathcal B}(\mathbf x_i)\) and local spectral energy \(E(\mathbf x_i)\) are computed at each window position \(\mathbf x_i\). To mitigate the impact of empty regions in sparse masks, the scale-specific global entropy is computed as an energy-weighted average:
\[
S_{\mathcal B}^{\text{scale}} = \frac{\sum_i s_{\mathcal B}(\mathbf x_i) E(\mathbf x_i)}{\sum_i E(\mathbf x_i)},
\]
where windows with negligible energy (\(E(\mathbf x_i) < 10^{-6}\)) are excluded. The final \(S_{\mathcal B}\) for an image is the mean over all defined scales. This multi-scale approach allows the metric to capture both fine-grained texture perturbations (at \(k=2\)) and macro-scale structural losses (at \(k=8\)) within a unified framework. \(\Delta S_{\mathcal B}\) is then computed as the difference between the multi-scale entropy of the masked image and that of the original fully sampled image.

\paragraph*{Metric comparison and visualization}
To validate the discriminative power of \(\Delta S_{\mathcal B}\) against standard metrics (PSNR and Spectral \(L_2\)), we quantify the \textbf{comparative advantage} of Random sampling over Periodic sampling. Since \(\Delta S_{\mathcal B}\) measures the distortion relative to the original signal, a lower absolute value indicates better information preservation. Therefore, for the comparison plots, we calculate the sample-wise performance gap defined as the \textit{reduction} in distortion provided by the Random geometry: specifically \(\Delta S_{\mathcal B}^{\text{periodic}} - \Delta S_{\mathcal B}^{\text{random}}\) for entropy, and the corresponding gain in PSNR or reduction in Spectral \(L_2\) error. These advantage scores are min-max normalized to allow simultaneous visualization on a common scale. We use violin plots with inner quartile lines (marking the 25th, 50th, and 75th percentiles) to explicitly depict the distribution of this advantage. 

\subsection{Wireless MIMO: simulation and over-the-air experiments}

\paragraph*{Channel model and simulation setup.}
We consider narrowband MIMO channels represented by complex matrices $\mathbf H\in\mathbb C^{N_r\times N_t}$, with $N_t$ transmit and $N_r$ receive antennas. Unless otherwise noted, we set $N_t=64$ and $N_r=16$ and assume uniform linear or planar arrays with half-wavelength antenna spacing. Channels are generated using a clustered geometric model with $L$ propagation paths, each characterized by a complex gain, angle of departure (AoD), and angle of arrival (AoA). Path gains are drawn from a complex Gaussian distribution with exponential power delay profile; AoAs/AoDs are drawn from Laplacian or Gaussian distributions around a few cluster centers, following standard massive MIMO models \cite{Marzetta2010MassiveMIMO, Jose2011PilotContam, Larsson2014CommMag}. Additive white Gaussian noise is added to the received pilots to achieve a target SNR (15 dB in the main experiments).

\paragraph*{Subsampling patterns and pilot budget}
Under a fixed pilot budget, only a subset $\Omega$ of entries of $\mathbf H$ is observed. We implement two acquisition geometries under matched budgets:
\begin{itemize}
    \item \textbf{Periodic deactivation:} we use a periodic deactivation pattern with period $d$: indices $i\equiv0\ (\text{mod}\  d)$ are turned off, and all remaining indices are observed (the complement set).   
    \item \textbf{Random deactivation:} the same number of antenna indices is selected uniformly at random without replacement along each axis, producing an irregular sampling pattern without long-range periodicity.
\end{itemize}
For each configuration $(d,\text{geometry})$, we generate multiple channel realizations and train a reconstruction network.

\paragraph*{CSI reconstruction and evaluation}
We use a Transformer-based completion network that takes as input a matrix of the same size as $\mathbf H$ where unobserved entries are set to zero together with a binary mask (or an equivalent two-channel representation). The network outputs a complex-valued estimate $\widehat{\mathbf H}$; real and imaginary parts are modeled as separate channels. The loss is the mean squared error between $\mathbf H$ and $\widehat{\mathbf H}$, and we report normalized mean squared error (NMSE) on held-out channels.

\paragraph*{Phase-space entropy for MIMO channels}
To compute $S_{\mathcal B}[\mathbf H]$, we apply the Husimi procedure to the magnitude matrix $|\mathbf H|$ (normalized to unit maximum per realization). We use a small 2D Gaussian window of size $\texttt{win}=4$, standard deviation $\sigma=1$, and hop size $\texttt{hop}=1$, which resolves local transitions between neighboring antennas while limiting spectral leakage. For each channel and subsampling pattern, we compute $\Delta S_{\mathcal B}$ by comparing the subsampled channel (with zeros at unobserved entries) to the fully sampled reference. These parameters are chosen to reflect the physical scale of local array structure.

\paragraph*{Over-the-air testbed}
To validate the simulation results in hardware, we perform over-the-air (OTA) experiments using a $1\times 8$ uniform linear array (ULA) at the transmitter and a $1\times 8$ ULA at the receiver (forming an $8\times 8$ MIMO link), two software-defined radios (SDRs), and an indoor line-of-sight environment (see SI Appendix for full details). One SDR acts as the transmitter, emitting pilot symbols over selected antennas according to the chosen subsampling pattern; the other acts as the receiver, measuring the responses across the $8$ receive antennas for each of the $8$ transmit antennas. The system operates at a carrier frequency of 1.2 GHz with a sampling rate of 5.6 MHz; transmit and receive gains are set to achieve an SNR of approximately 15 dB at the receiver input. The transmitter and receiver use internal oscillators without a shared reference clock, inducing realistic carrier frequency and sampling rate offsets. Measured CSI matrices are calibrated, subsampled according to Periodic or Random patterns, and fed to the same reconstruction network architecture as in simulation. $\Delta S_{\mathcal B}$ is computed from the measured $|\mathbf H|$ before training.

\subsection{MRI: accelerated multi-coil reconstruction and entropy auditing}

\paragraph*{Data and preprocessing}
We conduct experiments on complex-valued, fully sampled multi-coil MRI data. For each dataset, complex $k$-space measurements are provided per coil; we arrange them into arrays where the last dimension corresponds to the phase-encoding axis. Fully sampled reference images are obtained by applying a centered, orthonormal 2D inverse Fourier transform to the multi-coil $k$-space data, followed by root-sum-of-squares (RSS) combination across coils. All images are cropped or padded to a common size and normalized to $[0,1]$ by dividing by the maximum magnitude of the fully sampled RSS image.

\paragraph*{$k$-space undersampling masks}
To emulate accelerated acquisition, we generate one-dimensional binary masks along the phase-encoding axis (say, $k_y$) with a fixed central auto-calibration signal (ACS) region (e.g., 24 contiguous lines) always fully sampled. Outside the ACS region, we consider six mask families:
\begin{itemize}
    \item \textbf{Periodic:} phase-encoding lines are kept with a regular spacing determined by the target acceleration factor.
    \item \textbf{Random:} the required number of lines is selected uniformly at random outside the ACS region.
    \item \textbf{Poisson-disc--like variable density:} lines are drawn according to a variable density that decays with distance from the center, similar to \cite{Lustig2007SparseMRI}.
    \item \textbf{Three parametric variable-density masks:} probability distributions of the form
    \[
       p(k_y) \propto \bigl(1 + \alpha |k_y - k_{\mathrm{center}}|\bigr)^{-\beta},
    \]
    with $(\alpha,\beta)$ chosen in a pre-training manner by optimizing (i) the average $|\Delta S_{\mathcal B}|$ between zero-filled and fully sampled images, (ii) a $k$-space $L_2$ discrepancy, or (iii) pre-reconstruction PSNR, respectively, on the training set.
\end{itemize}
For each family and acceleration factor, a distinct mask is instantiated slice-wise (with the same ACS region) and used consistently for training and evaluation.

\paragraph*{Reconstruction network and metrics}
For each mask family and acceleration factor, we train a 2D U-Net \cite{Ronneberger2015UNet} that takes as input the zero-filled RSS image and outputs a reconstructed magnitude image. The U-Net has four encoding and four decoding stages with skip connections; convolutional layers use $3\times3$ kernels and ReLU activations. We train with an $\ell_1$ loss between reconstructed and fully sampled images using Adam (learning rate $2\times10^{-4}$, batch size 128) for 20 epochs. Reconstruction quality is evaluated on held-out slices using PSNR and structural similarity index (SSIM), computed slice-wise and averaged.

\paragraph*{Phase-space entropy for MRI}
We compute $\Delta S_{\mathcal B}$ for MRI on the magnitude images, comparing the zero-filled inputs to the fully sampled references. The Husimi parameters are set to $(\texttt{win},\sigma,\texttt{hop})=(32, 16, 10)$, so that each window spans a moderate fraction of the image. Within each window, we compute a Fourier transform, normalize the local power spectrum within the Nyquist band to obtain $\NB{\rho}$, and evaluate the Shannon entropy; $S_{\mathcal B}$ is the average over all windows. For all mask families and accelerations, we find $\Delta S_{\mathcal B}<0$, reflecting entropy loss due to missing spectral mass. We therefore focus on the magnitude $|\Delta S_{\mathcal B}|$ as a measure of how strongly the acquisition perturbs the original spectral organization.

\subsection{Parameter sensitivity and multi-scale robustness}
While the multi-scale energy-weighted integration (employed in the vision experiments) offers an automated solution to mitigate scale dependence, we also provide systematic ablations in the SI Appendix to validate the stability of \emph{single-scale} implementations. We find that the qualitative ranking of acquisition geometries (e.g., the advantage of Random over Periodic) remains invariant across a broad range of $(\texttt{win},\sigma,\texttt{hop})$ choices, provided these parameters do not fundamentally mismatch the signal's physical feature size. This confirms that while multi-scale aggregation is ideal for scale-variant data like natural images, a single instrument-matched scale suffices for domains with defined physical resolutions (such as MRI coils or MIMO arrays), and the metric functions as a stable macroscopic order parameter without requiring brittle fine-tuning.

\section{Supporting Information Appendix (SI)}
Long-form derivations (with distributional conventions), ablations, and OTA hardware details are provided in the SI Appendix.

\section{Significancestatement}
As machine learning is deployed across domains, practitioners still lack a common, physics-based way to judge whether as-acquired signals are ``good enough'' before committing to a model or training pipeline. We show that the acquisition-induced change in instrument-resolved phase-space entropy is a single, physics-based scalar that \emph{reflects downstream learnability}, predicting reconstruction and recognition difficulty from sampling geometry. This makes phase-space entropy a practical, model-free design knob for optimizing sensing policies and a common language for comparing acquisition quality across modalities.


\section{Declarations}
\textbf{Competing interests}: All authors declare no competing interests. \\
\textbf{Data availability}: The data used are all open source and can be found publicly. The MRI data used in this study are from the publicly available fastMRI multi-coil brain dataset \cite{Zbontar2018fastMRI}, available at https://fastmri.org/. \\
\textbf{Code availability}: Code for Husimi-based entropy auditing and mask generators can be found at https://doi.org/10.57760/sciencedb.31119 and https://doi.org/10.57760/sciencedb.31085. \\
\textbf{Author contribution}: 
conceptualization: Jun-Jie Zhang, Deyu Meng;
theory: Jun-Jie Zhang, Long-Gang Pang; 
experiments: Xiu-Cheng Wang, Jun-Jie Zhang, Nan Cheng;
writing: Jun-Jie Zhang, Xiu-Cheng Wang, Taijiao Du;
supervision: Deyu Meng.

\bibliography{pnas-sample.bib}

@article{Alkhateeb2014JSTSP,
  author  = {Ahmed Alkhateeb and Geert Leus and Robert W. Heath},
  title   = {Channel Estimation and Hybrid Precoding for Millimeter Wave Cellular Systems},
  journal = {IEEE Journal of Selected Topics in Signal Processing},
  year    = {2014},
  volume  = {8},
  number  = {5},
  pages   = {831--846},
  doi     = {10.1109/JSTSP.2014.2334278}
}

@inproceedings{Bao2022BEiT,
  author    = {Hangbo Bao and Li Dong and Songhao Piao and Furu Wei},
  title     = {BEiT: {BERT} Pre-Training of Image Transformers},
  booktitle = {International Conference on Learning Representations (ICLR)},
  year      = {2022},
  note      = {arXiv:2106.08254}
}

@inproceedings{Dosovitskiy2021An,
  title     = {An Image is Worth 16x16 Words: Transformers for Image Recognition at Scale},
  author    = {Alexey Dosovitskiy and Lucas Beyer and Alexander Kolesnikov and Dirk Weissenborn and Xiaohua Zhai and Thomas Unterthiner and Mostafa Dehghani and Matthias Minderer and Georg Heigold and Sylvain Gelly and Jakob Uszkoreit and Neil Houlsby},
  booktitle = {International Conference on Learning Representations (ICLR)},
  year      = {2021}
}

@inproceedings{Vinyals2016Matching,
  title     = {Matching Networks for One Shot Learning},
  author    = {Oriol Vinyals and Charles Blundell and Timothy Lillicrap and Koray Kavukcuoglu and Daan Wierstra},
  booktitle = {Advances in Neural Information Processing Systems (NeurIPS)},
  volume    = {29},
  year      = {2016},
  pages     = {3630--3638}
}

@article{CandesRombergTao2006Stable,
  author  = {Emmanuel J. Cand{\`e}s and Justin K. Romberg and Terence Tao},
  title   = {Stable Signal Recovery from Incomplete and Inaccurate Measurements},
  journal = {Communications on Pure and Applied Mathematics},
  year    = {2006},
  volume  = {59},
  number  = {8},
  pages   = {1207--1223},
  doi     = {10.1002/cpa.20124}
}

@article{CandesTao2006RUP,
  author  = {Emmanuel J. Cand{\`e}s and Terence Tao},
  title   = {Near-Optimal Signal Recovery From Random Projections: Universal Encoding Strategies?},
  journal = {IEEE Transactions on Information Theory},
  year    = {2006},
  volume  = {52},
  number  = {12},
  pages   = {5406--5425},
  doi     = {10.1109/TIT.2006.885507}
}

@article{CandesWakin2008IntroCS,
  author  = {Emmanuel J. Cand{\`e}s and Michael B. Wakin},
  title   = {An Introduction to Compressive Sampling},
  journal = {IEEE Signal Processing Magazine},
  year    = {2008},
  volume  = {25},
  number  = {2},
  pages   = {21--30},
  doi     = {10.1109/MSP.2007.914731}
}

@article{Cohen1989,
  author  = {Leon Cohen},
  title   = {Time--Frequency Distributions---A Review},
  journal = {Proceedings of the IEEE},
  year    = {1989},
  volume  = {77},
  number  = {7},
  pages   = {941--981},
  doi     = {10.1109/5.30749}
}

@book{Cover2006InfoTheory,
  author    = {Thomas M. Cover and Joy A. Thomas},
  title     = {Elements of Information Theory},
  edition   = {2nd},
  publisher = {John Wiley \& Sons},
  address   = {Hoboken, NJ},
  year      = {2006},
  series    = {Wiley Series in Telecommunications and Signal Processing},
  isbn      = {978-0-471-24195-9},
  doi       = {10.1002/047174882X},
  url       = {https://doi.org/10.1002/047174882X},
  note      = {First published 1991; Second edition includes new chapters on network information theory and rate-distortion theory}
}

@article{Zhang2025iScience,
  author  = {Jun-Jie Zhang and Dong-Xiao Zhang and Jian-Nan Chen and Long-Gang Pang and Deyu Meng},
  title   = {On the Uncertainty Principle of Neural Networks},
  journal = {iScience},
  year    = {2025},
  volume  = {28},
  number  = {4},
  pages   = {112197},
  issn    = {2589-0042},
  doi     = {10.1016/j.isci.2025.112197},
  url     = {https://doi.org/10.1016/j.isci.2025.112197}
}

@article{Donoho2006CS,
  author  = {David L. Donoho},
  title   = {Compressed Sensing},
  journal = {IEEE Transactions on Information Theory},
  year    = {2006},
  volume  = {52},
  number  = {4},
  pages   = {1289--1306},
  doi     = {10.1109/TIT.2006.871582}
}

@book{Flandrin1999,
  author    = {Patrick Flandrin},
  title     = {Time-Frequency/Time-Scale Analysis},
  publisher = {Academic Press},
  address   = {San Diego},
  year      = {1999},
  isbn      = {978-0-12-259870-8}
}

@article{Gao2016JSAC,
  author  = {Zhengshi Gao and Linglong Dai and Zhaocheng Wang and Sheng Chen and Zhaoyang Wang},
  title   = {Spatially Common Sparsity Based Adaptive Channel Estimation and Feedback for {FDD} Massive {MIMO}},
  journal = {IEEE Journal on Selected Areas in Communications},
  year    = {2016},
  volume  = {34},
  number  = {4},
  pages   = {998--1012},
  doi     = {10.1109/JSAC.2016.2545418}
}

@book{Goodfellow2016DeepLearning,
  author    = {Ian Goodfellow and Yoshua Bengio and Aaron Courville},
  title     = {Deep Learning},
  year      = {2016},
  publisher = {MIT Press},
  address   = {Cambridge, MA},
  series    = {Adaptive Computation and Machine Learning},
  edition   = {1},
  isbn      = {978-0262035613},
  url       = {http://www.deeplearningbook.org}
}

@inproceedings{He2022MAE,
  author    = {Kaiming He and Xinlei Chen and Saining Xie and Yanghao Li and Piotr Doll{\'a}r and Ross Girshick},
  title     = {Masked Autoencoders Are Scalable Vision Learners},
  booktitle = {Proceedings of the IEEE/CVF Conference on Computer Vision and Pattern Recognition (CVPR)},
  year      = {2022},
  pages     = {16000--16009}
}

@article{Husimi1940,
  author  = {K\={o}di Husimi},
  title   = {Some Formal Properties of the Density Matrix},
  journal = {Proceedings of the Physico-Mathematical Society of Japan. 3rd Series},
  year    = {1940},
  volume  = {22},
  pages   = {264--314}
}

@article{Jose2011PilotContam,
  author  = {J. Jose and A. Ashikhmin and T. L. Marzetta and S. Vishwanath},
  title   = {Pilot Contamination and Precoding in Multi-Cell {TDD} Systems},
  journal = {IEEE Transactions on Wireless Communications},
  year    = {2011},
  volume  = {10},
  number  = {8},
  pages   = {2640--2651},
  doi     = {10.1109/TWC.2011.060711.101155}
}

@book{KakSlaney2001CT,
  author    = {Avinash C. Kak and Malcolm Slaney},
  title     = {Principles of Computerized Tomographic Imaging},
  publisher = {SIAM},
  address   = {Philadelphia},
  year      = {2001},
  doi       = {10.1137/1.9780898719277}
}

@inproceedings{Krause2008Submodular,
  author    = {Andreas Krause and Ajit Singh and Carlos Guestrin},
  title     = {Near-Optimal Sensor Placements in Gaussian Processes: Theory, Efficient Algorithms and Empirical Studies},
  booktitle = {Journal of Machine Learning Research},
  volume    = {9},
  pages     = {235--284},
  year      = {2008}
}

@article{Larsson2014CommMag,
  author  = {Erik G. Larsson and Ove Edfors and Fredrik Tufvesson and Thomas L. Marzetta},
  title   = {Massive {MIMO} for Next Generation Wireless Systems},
  journal = {IEEE Communications Magazine},
  year    = {2014},
  volume  = {52},
  number  = {2},
  pages   = {186--195},
  doi     = {10.1109/MCOM.2014.6736761}
}

@article{LeCun2015DeepLearning,
  author  = {Yann LeCun and Yoshua Bengio and Geoffrey Hinton},
  title   = {Deep Learning},
  journal = {Nature},
  volume  = {521},
  number  = {7553},
  pages   = {436--444},
  year    = {2015},
  doi     = {10.1038/nature14539}
}

@article{Lustig2007SparseMRI,
  author  = {Michael Lustig and David L. Donoho and John M. Pauly},
  title   = {Sparse {MRI}: The Application of Compressed Sensing for Rapid {MR} Imaging},
  journal = {Magnetic Resonance in Medicine},
  year    = {2007},
  volume  = {58},
  number  = {6},
  pages   = {1182--1195},
  doi     = {10.1002/mrm.21391}
}

@book{Mallat2009Wavelet,
  author    = {St{\'e}phane Mallat},
  title     = {A Wavelet Tour of Signal Processing: The Sparse Way},
  edition   = {3rd},
  publisher = {Academic Press},
  address   = {Burlington, MA},
  year      = {2009},
  isbn      = {978-0123743701}
}

@article{Marzetta2010MassiveMIMO,
  author  = {Thomas L. Marzetta},
  title   = {Noncooperative Cellular Wireless with Unlimited Numbers of Base Station Antennas},
  journal = {IEEE Transactions on Wireless Communications},
  year    = {2010},
  volume  = {9},
  number  = {11},
  pages   = {3590--3600},
  doi     = {10.1109/TWC.2010.092810.091092}
}

@article{Ng2021DataCentricAI,
  author       = {Andrew Ng},
  title        = {A Chat with Andrew on {MLOps}: From Model-centric to Data-centric {AI}},
  year         = {2021},
  howpublished = {\url{https://www.deeplearning.ai/the-batch/a-chat-with-andrew-on-mlops-from-model-centric-to-data-centric-ai/}},
  note         = {Accessed: 2024-10-20}
}

@inproceedings{Pathak2016ContextEncoders,
  author    = {Deepak Pathak and Philipp Kr{\"a}henb{\"u}hl and Jeff Donahue and Trevor Darrell and Alexei A. Efros},
  title     = {Context Encoders: Feature Learning by Inpainting},
  booktitle = {Proceedings of the IEEE Conference on Computer Vision and Pattern Recognition (CVPR)},
  year      = {2016},
  pages     = {2536--2544},
  doi       = {10.1109/CVPR.2016.278}
}

@article{RechtFazelParrilo2010,
  author  = {Benjamin Recht and Maryam Fazel and Pablo A. Parrilo},
  title   = {Guaranteed Minimum-Rank Solutions of Linear Matrix Equations via Nuclear Norm Minimization},
  journal = {SIAM Review},
  year    = {2010},
  volume  = {52},
  number  = {3},
  pages   = {471--501},
  doi     = {10.1137/070697835}
}

@inproceedings{Ronneberger2015UNet,
  title     = {U-Net: Convolutional Networks for Biomedical Image Segmentation},
  author    = {Olaf Ronneberger and Philipp Fischer and Thomas Brox},
  booktitle = {International Conference on Medical Image Computing and Computer-Assisted Intervention (MICCAI)},
  pages     = {234--241},
  year      = {2015},
  organization = {Springer},
  note      = {LNCS 9351, doi:10.1007/978-3-319-24574-4\_28}
}

@book{Schleich2011QuantumOptics,
  author    = {Wolfgang P. Schleich},
  title     = {Quantum Optics in Phase Space},
  year      = {2011},
  publisher = {Wiley-VCH},
  address   = {Weinheim, Germany},
  isbn      = {978-3-527-40392-7},
  doi       = {10.1002/9783527635004},
  note      = {\url{https://doi.org/10.1002/9783527635004}}
}

@article{Shannon1948,
  author  = {Claude E. Shannon},
  title   = {A Mathematical Theory of Communication},
  journal = {Bell System Technical Journal},
  year    = {1948},
  volume  = {27},
  number  = {3},
  pages   = {379--423}
}

@article{Tropp2015RandMatrices,
  author  = {Joel A. Tropp},
  title   = {An Introduction to Matrix Concentration Inequalities},
  journal = {Foundations and Trends in Machine Learning},
  year    = {2015},
  volume  = {8},
  number  = {1--2},
  pages   = {1--230},
  doi     = {10.1561/2200000048}
}

@article{Wehrl1978,
  author  = {A. Wehrl},
  title   = {General Properties of Entropy},
  journal = {Reviews of Modern Physics},
  volume  = {50},
  number  = {2},
  pages   = {221--260},
  year    = {1978}
}

@article{Wigner1932,
  author  = {Eugene P. Wigner},
  title   = {On the Quantum Correction for Thermodynamic Equilibrium},
  journal = {Physical Review},
  year    = {1932},
  volume  = {40},
  number  = {5},
  pages   = {749--759},
  doi     = {10.1103/PhysRev.40.749}
}

@inproceedings{Xie2022SimMIM,
  author    = {Zhenda Xie and Zheng Zhang and Yue Cao and Yutong Lin and Jianmin Bao and Zhuliang Yao and Qiang Dai and Han Hu},
  title     = {SimMIM: A Simple Framework for Masked Image Modeling},
  booktitle = {Proceedings of the IEEE/CVF Conference on Computer Vision and Pattern Recognition (CVPR)},
  year      = {2022},
  pages     = {9653--9663}
}

@article{Zha2023DataCentricAI,
  author  = {Hongyuan Zha and Yujia He and Geng Liu and Jiachen Li and Di Wang and Hongyang Wang and Zhangyang Chen and Xiaolin Liu and Yang You and Cho-Jui Hsieh and Ce Zhang},
  title   = {Data-centric {AI}: A New Paradigm},
  journal = {arXiv preprint arXiv:2303.10158},
  year    = {2023}
}

@article{Zurek1994,
  author  = {Wojciech H. Zurek and Juan Pablo Paz},
  title   = {Decoherence, Chaos, and the Second Law},
  journal = {Physical Review Letters},
  volume  = {72},
  number  = {16},
  pages   = {2508--2511},
  year    = {1994}
}

@book{Arnold1989ClassicalMech,
  title     = {Mathematical Methods of Classical Mechanics},
  author    = {Vladimir I. Arnold},
  edition   = {2},
  year      = {1989},
  publisher = {Springer},
  address   = {New York}
}

@inproceedings{KrauseGuestrin2005GP,
  title     = {Near-Optimal Sensor Placements in {G}aussian Processes: Theory, Efficient Algorithms and Empirical Studies},
  author    = {Andreas Krause and Carlos Guestrin},
  booktitle = {Proceedings of the 22nd International Conference on Machine Learning (ICML)},
  year      = {2005}
}

@book{Pukelsheim2006OptimumDesign,
  author    = {Friedrich Pukelsheim},
  title     = {Optimal Design of Experiments},
  year      = {2006},
  publisher = {SIAM},
  address   = {Philadelphia, PA},
  series    = {Classics in Applied Mathematics},
  number    = {50},
  isbn      = {978-0-89871-604-8},
  doi       = {10.1137/1.9780898719100}
}

@book{Atkinson2007OptimumDesign,
  author    = {Anthony C. Atkinson and Alexander N. Donev and Randall D. Tobias},
  title     = {Optimum Experimental Designs, with SAS},
  series    = {Oxford Statistical Science Series},
  volume    = {34},
  publisher = {Oxford University Press},
  address   = {Oxford; New York},
  year      = {2007},
  pages     = {528},
  isbn      = {978-0-19-929659-0 (hardback) / 978-0-19-929660-6 (paperback)},
  note      = {Includes SAS program code and companion website}
}

@inproceedings{KohLiang2017Influence,
  title     = {Understanding Black-box Predictions via Influence Functions},
  author    = {Pang Wei Koh and Percy Liang},
  booktitle = {Proceedings of the 34th International Conference on Machine Learning (ICML)},
  year      = {2017},
  pages     = {1885--1894}
}

@inproceedings{Ghorbani2019DataShapley,
  title     = {Data Shapley: Towards Data Valuation using Shapley Value},
  author    = {Amirata Ghorbani and James Zou},
  booktitle = {Advances in Neural Information Processing Systems (NeurIPS)},
  year      = {2019},
  note      = {Also in ICML 2019 (PMLR 97: {Data Shapley: Equitable Valuation of Data for Machine Learning})}
}

@inproceedings{Mirzasoleiman2020CRAIG,
  title     = {Coresets for Data-efficient Training of Machine Learning Models},
  author    = {Baharan Mirzasoleiman and Jeff Bilmes and Jure Leskovec},
  booktitle = {Proceedings of the 37th International Conference on Machine Learning (ICML)},
  year      = {2020},
  pages     = {6950--6960}
}

@inproceedings{Killamsetty2020GLISTER,
  title     = {GLISTER: Generalization based Data Subset Selection for Efficient and Robust Learning},
  author    = {Krishnateja Killamsetty and Durga Sivasubramanian and Ganesh Ramakrishnan and Abir De and Rishabh Iyer},
  booktitle = {Proceedings of the AAAI Conference on Artificial Intelligence (AAAI)},
  year      = {2021}
}

@inproceedings{Killamsetty2021GradMatch,
  title     = {GRAD-MATCH: Gradient Matching based Data Subset Selection for Efficient Deep Model Training},
  author    = {Krishnateja Killamsetty and Durga Sivasubramanian and Ganesh Ramakrishnan and Abir De and Rishabh Iyer},
  booktitle = {Proceedings of the 38th International Conference on Machine Learning (ICML)},
  year      = {2021}
}

@inproceedings{Nguyen2020LEEP,
  title     = {{LEEP}: A New Measure to Evaluate Transferability of Learned Representations},
  author    = {Cuong V. Nguyen and Tal Hassner and Matthias Seeger and C{\'e}dric Archambeau},
  booktitle = {Proceedings of the 37th International Conference on Machine Learning (ICML)},
  year      = {2020},
  pages     = {7294--7305}
}

@inproceedings{You2021LogME,
  title     = {{LogME}: Practical Assessment of Pre-trained Models for Transfer Learning},
  author    = {Kaichao You and Yong Liu and Jianmin Wang and Mingsheng Long},
  booktitle = {Proceedings of the 38th International Conference on Machine Learning (ICML)},
  year      = {2021}
}

@article{10.1093/nsr/nwae141,
  author  = {Jun-Jie Zhang and Deyu Meng},
  title   = {Quantum-inspired Analysis of Neural Network Vulnerabilities: The Role of Conjugate Variables in System Attacks},
  journal = {National Science Review},
  pages   = {nwae141},
  year    = {2024},
  month   = {04},
  issn    = {2095-5138},
  doi     = {10.1093/nsr/nwae141},
  url     = {https://doi.org/10.1093/nsr/nwae141}
}

@article{Zbontar2018fastMRI,
  title={fastMRI: An Open Dataset and Benchmarks for Accelerated MRI},
  author={Zbontar, Jure and others},
  journal={arXiv preprint arXiv:1811.08839},
  year={2018}
}

\end{document}


\title{Supporting Information for\\
\textbf{Phase-space entropy at acquisition reflects downstream learnability}}

\author[2,3]{\fnm{Xiu-Cheng} \sur{Wang}}\email{xcwang\_1@stu.xidian.edu.cn}
\equalcont{Both authors contributed equally to this work.}

\author*[1]{\fnm{Jun-Jie} \sur{Zhang}}\email{zjacob@mail.ustc.edu.cn}
\equalcont{Both authors contributed equally to this work.}

\author[2,3]{\fnm{Nan} \sur{Cheng}}\email{dr.nan.cheng@ieee.org}

\author[4]{\fnm{Long-Gang} \sur{Pang}}\email{lgpang@ccnu.edu.cn}

\author[1]{\fnm{Taijiao} \sur{Du}}\email{dutaijiao@nint.ac.cn}

\author[5,6]{\fnm{Deyu} \sur{Meng}}\email{dymeng@mail.xjtu.edu.cn}

\affil[1]{\orgname{Northwest Institute of Nuclear Technology}, \orgaddress{\street{No. 28 Pingyu Road}, \city{Xi'an}, \postcode{710024}, \state{Shaanxi}, \country{China}}}

\affil[2]{\orgdiv{School of Telecommunications Engineering}, \orgname{Xidian University}, \orgaddress{\street{No. 2 South Taibai Road}, \city{Xi'an}, \postcode{710071}, \state{Shaanxi}, \country{China}}}
\affil[3]{\orgname{State Key Laboratory of ISN}, \orgaddress{\street{No. 2 South Taibai Road}, \city{Xi'an}, \postcode{710071}, \state{Shaanxi}, \country{China}}}

\affil[4]{\orgdiv{Key Laboratory of Quark and Lepton Physics (MOE) \& Institute of Particle Physics}, \orgname{Central China Normal University}, \orgaddress{\street{No. 152 Luoyu Road}, \city{Wuhan}, \postcode{430079}, \state{Hubei}, \country{China}}}

\affil[5]{\orgdiv{Ministry of Education Key Lab of Intelligent Networks and Network Security}, \orgname{Xi’an Jiaotong University}, \orgaddress{\street{No. 28 Xianning West Road}, \city{Xi'an}, \postcode{710049}, \state{Shaanxi}, \country{China}}}

\affil[6]{\orgdiv{School of Mathematics and Statistics}, \orgname{Xi’an Jiaotong University}, \orgaddress{\street{No. 28 Xianning West Road}, \city{Xi'an}, \postcode{710049}, \state{Shaanxi}, \country{China}}}

\maketitle

\bigskip
\noindent\textbf{This PDF includes}
\begin{enumerate}
  \item Supporting Note 1: Theoretical Derivations and Proofs for Optimality of Random Image Masking
  \item Supporting Note 2: MIMO Over-the-Air Hardware Configuration
  \item Supporting Note 3: Ablation Studies on Husimi Parameters (\texttt{win}, $\sigma$, \texttt{hop})
  \item Supporting Note 4: Quality–entropy correlation plots for MRI
\end{enumerate}

\clearpage
\section{Supporting Note 1: Theoretical Derivations and Proofs for Optimality of Random Image Masking}

\subsection{Physical Perspective: Symplectic Structure, Liouville’s Theorem, and Entropy}

Liouville’s theorem provides a motivating reference point: for an exactly transported density under an ideal reversible (symplectic) flow, phase-space volume is preserved and the \emph{unrestricted} Shannon entropy does not change. In this work we do not invoke a strict conservation law for our \emph{coarse-grained, band-limited} entropy; instead we use it as a practical proxy for acquisition-induced information mixing or removal at finite resolution.

The purpose of this work is to clarify how two common masking strategies affect phase-space entropy:

\begin{enumerate}
    \item \textbf{Periodic masking.} \\
    This operation introduces structured replicas of the spectrum in frequency space. High-frequency energy is folded (aliased) into the low-frequency region, creating irreversible mixing in phase space. By strict concavity and Jensen’s inequality, one can show that this aliasing always leads to entropy increase, and the increase is strict whenever overlapping replicas occur.

    \item \textbf{Random masking (random subsampling).} \\
    Unlike periodic masking, random subsampling does not maintain fixed lattice coherence. In the limit of a large number of degrees of freedom, its effect in phase space is equivalent to adding a small, unbiased, isotropic perturbation in frequency. Since high frequencies are no longer systematically folded into the low-frequency baseband, the phase-space entropy is preserved in expectation, with only small $O(N^{-1/2})$ fluctuations.
\end{enumerate}

\paragraph{Physical intuition.}
Periodic masking acts like a \emph{regular folding mechanism}: it systematically overlays high-frequency content onto the low-frequency region, causing irreversible information mixing and an increase in entropy. Random masking acts more like a \emph{uniform jitter}: it breaks up coherent folding patterns, spreading energy in an unbiased way that does not accumulate into systematic aliasing. As the image size grows, the average phase-space entropy remains nearly unchanged.

In summary, the analysis shows a sharp contrast: \textbf{periodic subsampling necessarily increases entropy (irreversible loss of information), while random subsampling approximately preserves entropy in the statistical limit (information is nearly conserved).}

\subsection{Stationarity in Natural Images}
Natural images are often modeled as approximately stationary random processes: their local statistics change slowly across space. Formally, for an image function \( I(x,y) \), we have
\begin{equation}
\mathbb{E}[I(x-a_x, y-a_y)] \approx \mathbb{E}[I(x,y)], \quad \forall (a_x,a_y)\in \mathbb{R}^2.
\end{equation}
Here, $\mathbb{E}[\cdot]$ denotes the ensemble average across samples; we assume weak ergodicity, so local spatial averaging approximates the ensemble expectation. This approximate stationarity motivates analyzing images in a joint phase-space representation, where spatial and spectral information coexist. Such a representation allows us to study how operations like masking affect both spatial detail and frequency content simultaneously.

\subsection{Wigner Transform and Phase Space Density}

To analyze the information entropy, we introduce the Wigner transform. For an image \(I(\mathbf{x}) \), where $\mathbf{x}=(x,y)$, the two-dimensional Wigner distribution is defined as:
\begin{equation}
W_I(\mathbf{x}; \mathbf{k}) = \int I\left(\mathbf{x}+\frac{\boldsymbol{\mathbf{\xi}}}{2}\right) I^*\left(\mathbf{x}-\frac{\boldsymbol{\mathbf{\xi}}}{2}\right) e^{-i\mathbf{k}\cdot\boldsymbol{\mathbf{\xi}}}  \dd\boldsymbol{\mathbf{\xi}}.
\end{equation}
The Wigner transform creates a phase-space representation of an image, combining spatial domain $\mathbf{x}$ and frequency domain $\mathbf{k}=(k_x, k_y)$. Natural images are not purely localized in space or frequency, so analyzing them in a joint domain avoids losing structural information. This makes it possible to evaluate how masking affects both spatial details and spectral content together. 

We adopt the symmetric Fourier convention where the Wigner product–convolution carries a prefactor $(2\pi)^{-2}$; all delta masses and normalization constants are stated compatibly with that convention.

Since \( W_I \) can take negative values, we choose a \emph{Cohen’s class} kernel that is
\emph{known} to produce a non-negative distribution (e.g., the
Husimi distribution, which correspond to convolving \(W_I\) with the Wigner of a window
function such as a minimum-uncertainty Gaussian). Define
\begin{equation}
\rho_I(X)\;=\;(\Phi*W_I)(X),\qquad
\Phi\ge 0,\quad X=(\mathbf{x},\mathbf{k}),
\end{equation}
where \(*\) denotes convolution in \((x,y,k_x,k_y)\). 
For convenience, we write “\(*\)” for convolution in full phase space \((\mathbf{x},\mathbf{k})\), and
“\(\ast_{\mathbb{T}^2}\)” for convolution over the frequency torus \(\mathbb{T}^2\) (in \(\mathbf{k}\) only). Throughout this paper, we use periodic boundary conditions in the frequency domain (\(\mathbb{T}^2\)) for the analysis, ensuring consistency across all frequency space representations. This allows for proper application of symplectic transformations and the preservation of phase-space volume.
\emph{Not every} Gaussian convolution guarantees non-negativity; here we
specifically use kernels (e.g., spectrogram/Husimi) that do.

We assume that the smoothing kernel \(\Phi\) may vary across different domains to reflect the specific physical resolution of each measurement device. In some cases, \(\Phi\) might represent the imaging resolution in MRI, or the antenna resolution in wireless MIMO systems.

\subsection{Entropy and Liouville Dynamics}

We quantify the spread of information in phase space using the Shannon entropy of an
\emph{unnormalized} smoothed Wigner density $\rho(X)$:
\begin{equation}
S[\rho]\;:=\;-\int \rho(X)\,\ln \rho(X)\,\dd X.
\end{equation}
When the image undergoes an ideal (canonical/symplectic) transformation, $\rho$ evolves by a
Liouville (continuity) equation
\begin{equation}\label{eq:liouville}
\partial_t \rho\;+\;\nabla_X \cdot (\rho\,V)\;=\;0,\qquad \nabla_X\!\cdot V=0,
\end{equation}
where $V$ is the phase-space flow field. Under \eqref{eq:liouville} and suitable boundary
conditions (torus or fast decay), phase-space volume is preserved and
\begin{equation}
\frac{\mathrm d}{\mathrm dt}\,S[\rho]\;=\;0.
\end{equation}

In our analyses we often use a \emph{band-normalized} density over the frequency torus
$\mathcal B\subset\mathbb{T}^2$ at each spatial location:
\begin{equation}
m[\rho](\mathbf{x}):=\int_{\mathcal B}\rho(\mathbf{x};\mathbf{k})\,\dd\mathbf{k},
\qquad
\NB{\rho}(\mathbf{x};\mathbf{k}):=\frac{\rho(\mathbf{x};\mathbf{k})}{m[\rho](\mathbf{x})}.
\end{equation}
The corresponding local band-entropy and a spatial aggregate are
\begin{equation}
s_{\mathcal B}[\NB{\rho}](\mathbf{x})
:=-\int_{\mathcal B}\NB{\rho}(\mathbf{x};\mathbf{k})\,
\ln \NB{\rho}(\mathbf{x};\mathbf{k})\,\dd\mathbf{k},
\qquad
S_{\mathcal B}[\NB{\rho}]
:=\int s_{\mathcal B}[\NB{\rho}](\mathbf{x})\,w(\mathbf{x})\,\dd\mathbf{x},
\end{equation}
where $w(\mathbf{x})$ is a specified weight (e.g., uniform or energy-weighted
$w\propto m[\rho](\mathbf{x})$). Since $\NB{\rho}$ includes an $\mathbf{x}$-dependent
normalization, exact Liouville conservation does not strictly apply to
$S_{\mathcal B}[\NB{\rho}]$; it \emph{serves as an ideal reference}
when smoothing and normalization are fixed, and becomes exact for $\rho$ under
\eqref{eq:liouville}.

\subsection{Effect of Masking on Phase Space Entropy}

A masking operation is defined as
\begin{equation}
J(\mathbf{x}) = M(\mathbf{x})\,I(\mathbf{x}),
\end{equation}
where \(M(\mathbf{x})\) is the mask and \(I(\mathbf{x})\) is the image. The bold vector $\mathbf{x}=(x,y)$. Applying the Wigner transform to the masked image \(J = M \cdot I\) yields
\begin{equation}
W_J(\mathbf{x},\mathbf{k}) = \int M\!\left(\mathbf{x}+\tfrac{\boldsymbol{\mathbf{\xi}}}{2}\right)M^*\!\left(\mathbf{x}-\tfrac{\boldsymbol{\mathbf{\xi}}}{2}\right)\,
            I\!\left(\mathbf{x}+\tfrac{\boldsymbol{\mathbf{\xi}}}{2}\right)I^*\!\left(\mathbf{x}-\tfrac{\boldsymbol{\mathbf{\xi}}}{2}\right)\,
            e^{-i\mathbf{k}\cdot\boldsymbol{\mathbf{\xi}}}\,\dd\boldsymbol{\mathbf{\xi}},
\end{equation}
where $\mathbf{k}=(k_x,k_y)$ and $\boldsymbol{\mathbf{\xi}}=(\xi_x,\xi_y)$.
We now proceed to express this in a more manageable form in 2D. Using the inverse-Fourier
representation of the Wigner distribution, we have
\begin{equation}
I\!\left(\mathbf{x}+\tfrac{\boldsymbol{\mathbf{\xi}}}{2}\right)I^*\!\left(\mathbf{x}-\tfrac{\boldsymbol{\mathbf{\xi}}}{2}\right)
=\frac{1}{(2\pi)^2}\int_{\mathbb{R}^2} W_I(\mathbf{x},\mathbf{k}')\,e^{i\,\mathbf{k}'\cdot\boldsymbol{\mathbf{\xi}}}\,\dd \mathbf{k}',
\end{equation}
and similarly
\begin{equation}
M\!\left(\mathbf{x}+\tfrac{\boldsymbol{\mathbf{\xi}}}{2}\right)M^*\!\left(\mathbf{x}-\tfrac{\boldsymbol{\mathbf{\xi}}}{2}\right)
=\frac{1}{(2\pi)^2}\int_{\mathbb{R}^2} W_M(\mathbf{x},\mathbf{q})\,e^{i\,\mathbf{q}\cdot\boldsymbol{\mathbf{\xi}}}\,\dd \mathbf{q}.
\end{equation}
Substituting into the definition of \(W_J\) and integrating over \(\boldsymbol{\mathbf{\xi}}\) gives the
product–convolution law in frequency:
\begin{equation}\label{eq:prod-conv}
W_J(\mathbf{x},\mathbf{k})
=\frac{1}{(2\pi)^2}\int_{\mathbb{R}^2} W_M(\mathbf{x},\mathbf{q})\,
W_I\big(\mathbf{x},\mathbf{k}-\mathbf{q}\big)\,\dd\mathbf{q}.
\end{equation}
The numerical prefactor \((2\pi)^{-2}\) follows the symmetric Fourier normalization stated above, 
ensuring internal consistency across all Wigner and convolution relations.

Let the observed (nonnegative) phase-space density be \(\rho=\Phi*W\) with a fixed
nonnegative Cohen-class kernel \(\Phi\) (convolution in both $\mathbf{x}$ and $\mathbf{k}$).
From the product–convolution law, \(\rho_J=\Phi*W_J=\frac{1}{(2\pi)^2}\,\Phi*\big(W_M \ast_{\mathbb{T}^2} W_I\big)\). 

After smoothing both the mask and the image with the kernel \(\Phi\), we write
\begin{equation}
\rho_J=\Phi*W_J,\qquad \rho_I=\Phi*W_I,\qquad K_M:=\Phi*W_M .
\end{equation}
Then, using the product–convolution law and the local-averaging approximation discussed above,
\begin{equation}\label{eq:phi-conv-approx}
\rho_J(\bm x,\bm k)\;\approx\;\frac{1}{(2\pi)^2}\,K_M \ast_{\mathbb{T}^2}\rho_I(\bm x,\bm k).
\end{equation}
The kernel \(K_M\) is nonnegative and plays the role of an effective kernel that modulates both spatial and frequency components of the image. This convolution reflects how masking affects the phase-space distribution and hence its entropy.

Unlike the Liouville flow, which preserves phase space volume (i.e., no entropy change), the masking operation is not a canonical transformation and thus can break entropy conservation. The extent and direction of entropy change depend on the structure of the mask and on how the smoothing operation suppresses oscillatory cross-terms in \(W_M\). In particular, periodic masks can introduce aliasing effects, leading to an increase in entropy. The smoothed phase-space density \(\rho_J(\mathbf{x},\mathbf{k})\) is a convex combination of shifted versions of \(\rho_I(\mathbf{x},\mathbf{k})\), and entropy is generally increased due to the mixing of spatial and frequency components in the distribution.

\subsubsection{Periodic Masking Causes Entropy Increase (Periodic Subsampling)}

\paragraph{Definition of a periodic mask.}
We model periodic subsampling by a 2D Dirac comb with spacings $d_x,d_y$:
\begin{equation}
M_{\text{per}}(x,y)\;=\;\sum_{m,n\in\mathbb Z}\delta(x-m d_x)\,\delta(y-n d_y).
\end{equation}
Within a large apodization window of area $A$ (later $A\!\to\!\infty$), this corresponds to keeping only pixels on a regular lattice and discarding all others. By the 2D Poisson summation formula (in the sense of tempered distributions),
\begin{equation}
M_{\text{per}}(x,y)
\;=\;\frac{1}{d_x d_y}\sum_{n\in\mathbb Z^2}
\exp\!\Big(i\,\kappa_n\!\cdot\!(x,y)\Big),\qquad
\kappa_n=\Big(\tfrac{2\pi n_x}{d_x},\,\tfrac{2\pi n_y}{d_y}\Big).
\end{equation}
Thus $M_{\text{per}}$ admits a Fourier series with \emph{equal} coefficients $c_n=\frac{1}{d_x d_y}$ for all $n\in\mathbb Z^2$.

\paragraph{Wigner distribution of the mask.}
Substitute the Fourier series of $M_{\text{per}}$ into the Wigner definition. Writing
\(
M_{\text{per}}(x,y)=\sum_{n} c_n e^{i \kappa_n\cdot(x,y)}
\)
with $c_n=\frac{1}{d_x d_y}$, the algebra is identical to the general periodic case, and yields
\begin{equation}
W_{M_{\text{per}}}(x,y;k_x,k_y)
=(2\pi)^2\!\!\sum_{n,m\in\mathbb Z^2} 
c_n c_m^*\,
\delta\!\left((k_x,k_y)-\tfrac{\kappa_n+\kappa_m}{2}\right)\,
e^{i(\kappa_n-\kappa_m)\cdot(x,y)}.
\end{equation}
The \emph{auto-terms} ($n=m$) are located at $k=\kappa_n$ and are independent of $(x,y)$; the \emph{cross-terms} ($n\neq m$) oscillate in $(x,y)$ with spatial frequencies $\kappa_n-\kappa_m$ and sit at midpoints $\tfrac{\kappa_n+\kappa_m}{2}$ in frequency.

\paragraph{Smoothing and the effective kernel.}
Let the observed (nonnegative) phase-space density be the Cohen-class smoothing of the Wigner distribution with a kernel $\Phi$:
\begin{equation}
\rho_J \;=\; \Phi * W_J,\qquad K_{M_{\text{per}}} \;:=\; \Phi * W_{M_{\text{per}}}.
\end{equation}
Because $\Phi$ averages over a few unit cells in $(x,y)$, the rapid oscillations $e^{i(\kappa_n-\kappa_m)\cdot(x,y)}$ for $n\!\neq\! m$ are suppressed by destructive averaging (Riemann–Lebesgue). Retaining the dominant auto-terms gives
\begin{equation}
K_{M_{\text{per}}}(k_x,k_y)
\;\approx\; (2\pi)^2 \sum_{n\in\mathbb Z^2} |c_n|^2\,
\big(\Phi\ast_{\mathbb{T}^2}\delta(\cdot-\kappa_n)\big)(k_x,k_y).
\end{equation}
Since $|c_n|^2=\tfrac{1}{d_x^2 d_y^2}$ is constant, $K_{M_{\text{per}}}$ is a \emph{broadened frequency comb} centered at the reciprocal-lattice points $\{\kappa_n\}$ with nonnegative weights proportional to $|c_n|^2$.

\paragraph{Resulting phase-space density via the product–convolution law.}
For a masked image $J=M_{\text{per}}\cdot I$, the Wigner transform of the product obeys (in 2D, cf. the standard derivation in 1D and extending component-wise)
\begin{equation}
W_J(x,y;k_x,k_y)\;=\;\frac{1}{(2\pi)^2}
\iint_{\mathbb R^2}\! W_{M_{\text{per}}}(x,y;q_x,q_y)\;
W_I\big(x,y;k_x-q_x,\,k_y-q_y\big)\,\dd\mathbf{q}.
\end{equation}
Applying the linear smoothing \(\Phi\) and using the same local-averaging approximation (under which \(\Phi\) effectively commutes with the frequency convolution),
\begin{equation}
\rho_J \;=\; \Phi*W_J 
\;=\; \frac{1}{(2\pi)^2}\,(\Phi*W_{M_{\text{per}}}) \ast_{\mathbb{T}^2} (\Phi*W_I)
\;=\; \frac{1}{(2\pi)^2}\, K_{M_{\text{per}}} \ast_{\mathbb{T}^2} \rho_I .
\end{equation}
On the frequency torus \(\mathbb{T}^2\) (i.e., within one Nyquist cell modulo wrap-around),
the \(\Phi\)-averaged periodic kernel \(K_{M_{\mathrm{per}}}\) is a broadened comb centered at the reciprocal-lattice points modulo the cell. Consequently,
\begin{equation}
\rho_J(x,y;\mathbf{k})\;\approx\;\frac{1}{(2\pi)^2}\,\big(K_{M_{\mathrm{per}}}\ast_{\mathbb{T}^2}\rho_I\big)(x,y;\mathbf{k}).
\end{equation}

Using that \(K_{M_{\mathrm{per}}}\) is (after spatial averaging) effectively independent of
\((x,y)\), we obtain the convex-mixture form
\begin{equation}
\NB{{\rho}_J}(\mathbf{x};\mathbf{k})
\;\approx\;\sum_{\mathbf{\kappa}_n\in\mathcal B} w_n\,
\NB{{\rho}_I}\big(\mathbf{x};\mathbf{k}-\mathbf{\kappa}_n\big),
\qquad
w_n=\frac{\int_{\mathcal B}K_{M_{\mathrm{per}}}(\mathbf{k}-\mathbf{\kappa}_n)\,\dd\mathbf{k}}
{\sum_{\mathbf{\kappa}_m\in\mathcal B}\int_{\mathcal B}K_{M_{\mathrm{per}}}(\mathbf{k}-\mathbf{\kappa}_m)\,\dd\mathbf{k}}.
\end{equation}
Here $w_n\ge 0$ and $\sum_n w_n=1$, and crucially $w_n$ do not depend on $\mathbf{k}$ (and,
after averaging, not on $\mathbf{x}$ either), so the right-hand side is a \emph{pointwise} convex
combination in $\mathbf{k}$ at each $\mathbf{x}$.

This formula shows that periodic subsampling replicates $\rho_I$ on the reciprocal lattice and superposes the copies. Physically, frequencies outside the baseband are shifted by $\kappa_n$ back into it—\emph{high-frequency energy is folded (aliased) into low frequency}, producing irreversible mixing.

\paragraph{Entropy increase via Jensen’s inequality.}
Let \( f(t)=-t\ln t \), which is strictly concave on \((0,1]\).
Since the normalized density after masking can be expressed as
\begin{equation}
\NB{\rho_J}(\mathbf{x};\mathbf{k})
=\sum_n w_n\,r_n(\mathbf{k}), \qquad
r_n(\mathbf{k}) = \NB{\rho_I}(\mathbf{x};\mathbf{k}-\mathbf{\kappa}_n),
\end{equation}
and \( \sum_n w_n=1 \), Jensen’s inequality yields
\begin{equation}
\int_{\mathcal B} f\!\Big(\NB{\rho_J}(\mathbf{x};\mathbf{k})\Big)\dd\mathbf{k}
\ge
\sum_n w_n \int_{\mathcal B} f\!\big(r_n(\mathbf{k})\big)\dd\mathbf{k}.
\end{equation}
That is,
\begin{equation}
s_{\mathcal B}[\NB{\rho_J}](\mathbf{x})
\ge
s_{\mathcal B}[\NB{\rho_I}](\mathbf{x}),
\end{equation}
with equality only when all \( r_n(\mathbf{k}) \) (for \( w_n>0 \)) are identical almost everywhere on \( \mathcal B \).

\subsubsection{Random Masking Approximately Preserves Entropy (Random Subsampling)}

\paragraph{Definition of a random subsampling mask.}
Let the underlying fine grid have pixel pitches $p_x,p_y$ and index set 
$\mathcal{G}=\{(m p_x, n p_y): m,n\in\mathbb Z\}$ within a large window of area $A$ (later $A\!\to\!\infty$).
Define an i.i.d.\ Bernoulli field\footnote{The theoretical analysis uses Bernoulli sampling because it yields closed-form expectations, whereas the experiments use uniform sampling without replacement under fixed budgets. For moderate budgets these two behave similarly in entropy.} $\{b_{m,n}\}$ with 
\begin{equation}
b_{m,n}\in\{0,1\},\qquad \mathbb{P}(b_{m,n}=1)=\rho\in(0,1),
\end{equation}
and set the (energy-normalized) random subsampling mask
\begin{equation}
M_{\mathrm{rand}}(x,y)\;=\;\frac{1}{\sqrt{\rho}}\sum_{(m,n)\in\mathbb Z^2}
b_{m,n}\,\delta(x-m p_x)\,\delta(y-n p_y).
\end{equation}
The global factor $1/\sqrt{\rho}$ keeps the average mask energy (Frobenius norm) independent of $\rho$, so we can compare entropies under a common normalization.\footnote{%
If one prefers unit amplitude on the kept pixels, i.e.\ $M=\sum b_{m,n}\delta(\cdot)$, one can perform unit-mass normalization on the final phase-space density $\rho_J$ instead, with equivalent results; this paper chooses energy normalization for direct comparison with the ``periodic mask’’ in terms of capacity and Fisher information.}

\paragraph{Wigner distribution of the mask: mean--fluctuation decomposition.}
Write $b_{m,n}=\rho+\epsilon_{m,n}$ with $\mathbb{E}[\epsilon_{m,n}]=0$,
$\mathrm{Var}(\epsilon_{m,n})=\rho(1-\rho)$ and independence across $(m,n)$.
Then
\begin{equation}
M_{\mathrm{rand}}\;=\;\underbrace{\sqrt{\rho}\,\Gamma}_{\text{mean part}}
\;+\;\underbrace{\frac{1}{\sqrt{\rho}}\sum \epsilon_{m,n}\,\delta(\cdot-mp_x)\delta(\cdot-np_y)}_{\text{zero-mean fluctuation}},
\end{equation}
where $\Gamma(x,y)=\sum_{m,n}\delta(x-m p_x)\delta(y-n p_y)$ is the full lattice comb.
By bilinearity, the Wigner distribution splits into
\begin{equation}
W_{M_{\mathrm{rand}}}
\;=\; \rho\,W_{\Gamma} \;+\; W_{\varepsilon}
\;+\; 2\,\mathrm{Re}\,W_{\Gamma,\varepsilon},
\end{equation}
where $W_{\varepsilon}$ is the Wigner of the fluctuation part and $W_{\Gamma,\varepsilon}$ the cross-Wigner.
Taking expectations over the mask randomness yields
\begin{equation}
\mathbb{E}\big[W_{M_{\mathrm{rand}}}\big]
\;=\; \rho\,W_{\Gamma} \;+\; \mathbb{E}\big[W_{\varepsilon}\big],
\quad\text{since}\quad \mathbb{E}\big[W_{\Gamma,\varepsilon}\big]=0
\end{equation}
(the cross term vanishes because one factor has zero mean).

We now analyze the two contributions:

\smallskip
\noindent\emph{(i) The mean term $\rho\,W_{\Gamma}$.}
As in the periodic case, $W_{\Gamma}$ consists of sharp peaks at the reciprocal lattice
$\kappa_{u,v}=(2\pi u/p_x,\,2\pi v/p_y)$ plus oscillatory cross-terms in $(x,y)$.
After smoothing in $(x,y)$ with kernel $\Phi$ over a few unit cells, the oscillatory cross-terms of $W_\Gamma$ are suppressed (Riemann--Lebesgue averaging), leaving broadened peaks centered at all reciprocal lattice points $\kappa_{u,v}$. 
However, if one restricts attention to the baseband detection region 
$|k_x|\le \pi/p_x,\;|k_y|\le \pi/p_y$, then only the central lobe at $\mathbf{k}=(0,0)$ remains inside the band.
\footnote{Under a strictly periodic mask the Wigner spectrum forms a comb on the reciprocal lattice. With Bernoulli coefficients, however, lattice coherence is destroyed in expectation: after spatial averaging by \(\Phi\), nonzero harmonics lose phase coherence so that only a DC spike at \(\mathbf{k}=\mathbf{0}\) remains coherent, while the residual appears as an approximately flat pedestal in \(k\). The pedestal’s contribution to the \emph{smoothed kernel} decays like \(O(1/N)\) due to averaging over \(N\) effectively independent spatial cells.}

\smallskip
\noindent\emph{(ii) The fluctuation term $\mathbb{E}[W_{\varepsilon}]$.}
By independence and zero-mean, $\mathbb{E}[\epsilon_{m_1,n_1}\epsilon_{m_2,n_2}]=0$ when $(m_1,n_1)\neq (m_2,n_2)$ and equals $\rho(1-\rho)$ on the diagonal.
Tracing the definition of $W$ shows that only the \emph{diagonal} pairs force $\xi\!=\!0$ and hence contribute a $k$-independent pedestal.
Thus $\mathbb{E}[W_{\varepsilon}]$ produces a \emph{flat (white) background in $k$}, localized in $(x,y)$ on the lattice and becoming spatially uniform after smoothing by $\Phi$ over many unit cells.

\paragraph{Smoothing and the effective kernel.}
Let \(K_{M_{\mathrm{rand}}}:=\Phi*W_{M_{\mathrm{rand}}}\).
Under the i.i.d.\ Bernoulli model and \(\Phi\)-averaging over many unit cells, we obtain
\begin{equation}
\mathbb{E}[K_{M_{\mathrm{rand}}}](\mathbf{k})
= (2\pi)^2\,\delta(\mathbf{k}) + \alpha_A\,\mathbf{1}_{\mathcal B}(\mathbf{k}),
\quad |\alpha_A|\le \frac{C}{N}.
\end{equation}

Here \(\delta(\mathbf{k})\) denotes the identity distribution on the frequency torus
\(\mathbb{T}^2\), satisfying
\((g\ast_{\mathbb{T}^2}\delta)(\mathbf{k}) = g(\mathbf{k})\) for any test function \(g\).
With this convention, the convolution law
\(\rho_J=(2\pi)^{-2}K_{M_{\mathrm{rand}}}\ast_{\mathbb{T}^2}\rho_I\)
implies that
\begin{equation}
\mathbb{E}[\rho_J] = \rho_I + O\!\Big(\frac{1}{N}\Big).
\end{equation}

Moreover, since \(K_{M_{\mathrm{rand}}}\) is an average of \(O(N)\)
independent bounded contributions, standard concentration yields
\begin{equation}
\mathbb{E}\,\big\|K_{M_{\mathrm{rand}}}-\mathbb{E}[K_{M_{\mathrm{rand}}}]\big\|_{L^1(\mathcal{B})}
= O\!\Big(\tfrac{1}{\sqrt{N}}\Big).
\end{equation}
Combining the mean bias $O(\tfrac{1}{N})$ with the fluctuations $O(\tfrac{1}{\sqrt{N}})$ yields
\begin{equation}
\mathbb{E}\,\big\|\rho_J - \rho_I\big\|_{L^1(\mathcal{B})}
\;=\; O\!\Big(\tfrac{1}{\sqrt{N}}\Big).
\end{equation}

\paragraph{Band-normalization stability.}
Let \(\NB{\rho}\) denote the band-normalized density on \(\mathcal{B}\),
i.e.,
\begin{equation}
\NB{\rho}(\bm x;\bm k)=
\frac{\rho(\bm x;\bm k)}{m[\rho](\bm x)},\qquad
m[\rho](\bm x):=\int_{\mathcal B}\rho(\bm x;\tilde{\bm k})\,\dd\tilde{\bm k}.
\end{equation}
Assume \(m[\rho_I](\bm x),m[\rho_J](\bm x)\ge m_{\min}>0\) almost everywhere.
Then for each spatial location \(\bm x\),
\begin{equation}
\big\|\NB{\rho_J}(\bm x;\cdot)-\NB{\rho_I}(\bm x;\cdot)\big\|_{L^1_{\mathbf{k}}(\mathcal B)}
\le
\frac{2}{m_{\min}}\,
\big\|\rho_J(\bm x;\cdot)-\rho_I(\bm x;\cdot)\big\|_{L^1_{\mathbf{k}}(\mathcal B)}.
\end{equation}
Integrating over \(\mathbf{x}\) and using the \(O(N^{-1/2})\) concentration bound above gives
\begin{equation}
\mathbb{E}\big\|\NB{\rho_J}-\NB{\rho_I}\big\|_{L^1_{\mathbf{x},\mathbf{k}}(\mathcal B)}
=O\!\Big(\tfrac{1}{\sqrt{N}}\Big).
\end{equation}
This shows that the normalization step is Lipschitz-stable with respect
to the \(L^1\) perturbation of \(\rho\).

\paragraph{Physical explanation.}
Periodic masking creates coherent spectral replicas on the reciprocal lattice,
so high-frequency components are deterministically folded into the baseband,
producing irreversible aliasing and entropy growth.
Random masking, on the other hand, destroys lattice phase coherence:
after spatial averaging by \(\Phi\),
its spectrum contains only a central DC lobe plus a weak isotropic pedestal,
which corresponds to unbiased white perturbations in frequency.
Hence, high-frequency energy is not systematically folded into the low-frequency
region, and the phase-space entropy remains statistically invariant up to
\(O(N^{-1/2})\) fluctuations.
From an information-theoretic perspective, random masking results in a measurement
operator with a near-isotropic singular-value spectrum and well-conditioned Fisher
information, whereas periodic masking collapses certain singular values and
creates spectral holes that reduce information capacity.

\subsection{Assumptions, scope, and possible counterexamples}
\label{sec:assumptions}

For clarity and to delimit the regime in which the arguments and asymptotic bounds below hold, we list here the explicit assumptions used in the derivations above, the finite--$N$ conditions, and several noteworthy edge cases where the conclusions may fail. Note, these assumptions introduced in the analytical treatment of periodic and randomized subsampling are used solely to make the resulting phase-space mixing explicit. They are not prerequisites of the entropy principle itself, which is defined for arbitrary signals and sampling operators.

\paragraph{A1. Fourier convention.}
We adopt the forward/inverse Fourier transform convention (2D)
\begin{equation}
\widehat f(\mathbf k)=\int_{\mathbb R^2} f(\mathbf x)\,e^{-i\mathbf k\cdot\mathbf x}\,\dd\mathbf x,
\qquad
f(\mathbf x)=\frac{1}{(2\pi)^2}\int_{\mathbb R^2}\widehat f(\mathbf k)\,e^{i\mathbf k\cdot\mathbf x}\,\dd\mathbf k.
\end{equation}
All $(2\pi)$ prefactors appearing in the Wigner product–convolution relations follow from this convention.

\paragraph{A2. Smooth, nonnegative Cohen kernel.}
The smoothing kernel $\Phi$ belongs to the Schwartz class (or is a compactly supported, sufficiently smooth averaging kernel), satisfies $\Phi\ge0$, and has an effective spatial averaging radius $\ell_\Phi$ that covers many sampling unit cells (quantitatively: $\ell_\Phi\gg d$, where $d$ denotes the lattice spacing of a periodic mask). This scale separation guarantees suppression of rapidly oscillatory cross-terms by the Riemann--Lebesgue lemma in the large-window limit.

\paragraph{A3. Mask models.}
\begin{itemize}
  \item \emph{Periodic mask.} Modeled as an ideal Dirac comb $M_{\rm per}(x)=\sum_{m,n}\delta(x-m d_x)\delta(y-n d_y)$ (or its apodized approximation on a large window of area $A$), with exact lattice coherence.
  \item \emph{Random mask.} Modeled as i.i.d.\ Bernoulli samples $b_{m,n}\in\{0,1\}$ with $\mathbb P(b_{m,n}=1)=\rho\in(0,1)$ and finite variance; the mask is energy--normalized as in the main text ($1/\sqrt\rho$ prefactor). Independence (or sufficiently fast decay of correlations) across lattice sites is required for the concentration bounds.
\end{itemize}

\paragraph{A4. Large-sample averaging and finite--$N$ bounds.}
Let $N$ denote the effective number of independent spatial cells averaged by $\Phi$ (or present in the apodization window). The principal stochastic statements are asymptotic in $N\to\infty$; for finite $N$ we obtain the explicit bounds stated in the text:
\begin{equation}
\mathbb E\|K_{M_{\rm rand}}-\mathbb E K_{M_{\rm rand}}\|_{L^1(\mathcal B)}=O\!\Big(\tfrac{1}{\sqrt N}\Big),
\qquad
\mathbb E\|\rho_J-\rho_I\|_{L^1(\mathcal B)}=O\!\Big(\tfrac{1}{\sqrt N}\Big),
\end{equation}
with the constants depending on the per-cell variance and the $L^\infty$ bounds of the contributing terms; standard Hoeffding/Bernstein inequalities justify these rates under A3.

\paragraph{A5. Local energy (band) lower bound.}
We assume the local band energy satisfies $m[\rho](\mathbf x)\ge m_{\min}>0$ a.e., ensuring band--normalization is Lipschitz and avoiding division by numerically vanishing energy in the $L^1$ stability estimates.

\paragraph{A6. Image regularity and ergodicity.}
The image ensemble has bounded energy and sufficiently regular Wigner distributions so that integrals and exchanges of limits used in the derivations are justified. We further assume weak ergodicity/stationarity at the scale of $\Phi$ so that local spatial averages represent ensemble expectations.

\bigskip
\noindent\textbf{Remarks on scope.} Under A1--A6 the periodic mask analysis yields a pointwise convex combination of band densities (hence Jensen applies) and the random mask analysis yields the stated expectation and concentration bounds. When any assumption above is violated (see next subsection), the stated conclusions can fail or require modification.

\subsubsection*{Possible counterexamples and limiting cases}
The following non-exhaustive list highlights situations in which the above conclusions may not hold, or where the finite--$N$ corrections may be large.

\begin{itemize}
  \item \textbf{Lattice-invariant signals.} If the original band-normalized densities satisfy
  $r_n(\mathbf k)=\NB{\rho_I}(\mathbf x;\mathbf k-\kappa_n)$ that are identical (a.e.) for all $\kappa_n$ with $w_n>0$, then Jensen's inequality is tight and periodic subsampling does \emph{not} increase the band entropy. Such signals are exceptional but conceptually important (they are ``invariant under reciprocal lattice shifts'').
  \item \textbf{Strongly correlated or structured masks.} If the random mask exhibits long-range correlations, clustering, or periodic substructure, the decomposition into a DC spike plus small flat pedestal may fail; the effective kernel $K_M$ can keep coherent sidelobes and therefore produce deterministic aliasing-like effects.
  \item \textbf{Insufficient smoothing ($\ell_\Phi$ too small).} If the smoothing kernel $\Phi$ averages over too few unit cells, oscillatory cross-terms in $W_M$ may not be suppressed and the broadened-comb approximation for $K_{M_{\rm per}}$ breaks down.
  \item \textbf{Very small $N$.} For small numbers of independent cells, the $O(1/\sqrt N)$ fluctuations can be non-negligible and may lead to observable entropy shifts.
  \item \textbf{Near-zero local energy ($m_{\min}\approx 0$).} When local band energy is vanishingly small the normalization step amplifies perturbations and the Lipschitz bound becomes uninformative.
  \item \textbf{Deterministic masks engineered to avoid aliasing.} Special deterministic sampling patterns (e.g., certain nonuniform sampling designs used in compressed sensing) can produce well-conditioned measurement operators without randomness; such constructions lie outside the Bernoulli model and must be analyzed case by case.
\end{itemize}

\section{Supporting Note 2: MIMO Over-the-Air Hardware Configuration}
To validate the real-time reconstruction performance and practical viability of the proposed Transformer-based CSI completion network, a hardware testbed was established for over-the-air (OTA) MIMO signal transmission and reception, simulating a realistic edge computing scenario. The experimental platform comprises two National Instruments USRP-2901 software-defined radios (SDRs) and a Raspberry Pi 4B deployed for inference processing. Specifically, both the transmitter and receiver are equipped with a $1 \times 8$ uniform linear array (ULA), establishing an $8 \times 8$ MIMO communication link to facilitate the acquisition of high-dimensional CSI matrices. The transmitter is interfaced with a host PC running MATLAB R2024a for baseband channel sounding signal generation and control. Conversely, the receiver captures the wireless signals and performs downconversion and digitization to produce complex in-phase and quadrature (I/Q) samples for channel estimation. The experiments were conducted in a line-of-sight (LOS) indoor environment with a carrier frequency of 1.2 GHz and a sampling rate of 5.6 MHz. The transmitter and receiver gains were set to 25 dB and 40 dB, respectively, to achieve an approximate signal-to-noise ratio (SNR) of 15 dB.

\section{Supporting Note 3: Ablation Studies on Husimi Parameters}

\subsection{MIMO}
To ascertain the robustness of the proposed entropy metric with respect to its analysis parameters, we conducted an ablation study for the MIMO channel scenario. We varied the window size while holding the Gaussian width ($\sigma=1$) and hop size (hop=1) constant. This parameterization is due to the fundamental structural differences between natural images and MIMO channel matrices. While images exhibit strong 2D spatial correlations, the pertinent correlations within a channel matrix exist primarily along one-dimensional axes, corresponding to the antenna arrays at the transmitter or receiver. Setting hop=1 ensures a high-resolution scan across the antenna elements without downsampling, thereby capturing the fine-grained local spectral variations essential for assessing channel quality. Concurrently, a small $\sigma=1$ effectively localizes the Gaussian window's influence, concentrating the analysis on a single antenna's data stream (i.e., a row or column) at each step. This approach forces the 2D analysis framework to approximate a series of 1D evaluations, thereby minimizing cross-dimensional interference from physically uncorrelated antenna data and preserving the physical integrity of the spectral entropy measurement for the MIMO channel.

Across the majority of configurations, the observed trends aligned closely with theoretical expectations: as the deactivation interval $d$ increases (fewer antennas are deactivated), the magnitude of the band-entropy change, $\Delta S_{\mathcal B}$, decreases, indicating a less significant perturbation of the local spectral structure of the channel response. Furthermore, for any fixed deactivation rate, the \textsf{Periodic} masking scheme consistently yields larger $\Delta S_{\mathcal B}$ values than the \textsf{Random} masking scheme. This finding suggests that periodic deactivation introduces greater spectral disorder into the channel estimate, while random deactivation statistically preserves local entropy. Collectively, these results demonstrate that the entropy-based analysis provides a robust characterization of spectral degradation under different deactivation strategies and is not overly sensitive to the selection of the window size.

\begin{figure}[H]
    \centering
    \includegraphics[width=\textwidth]{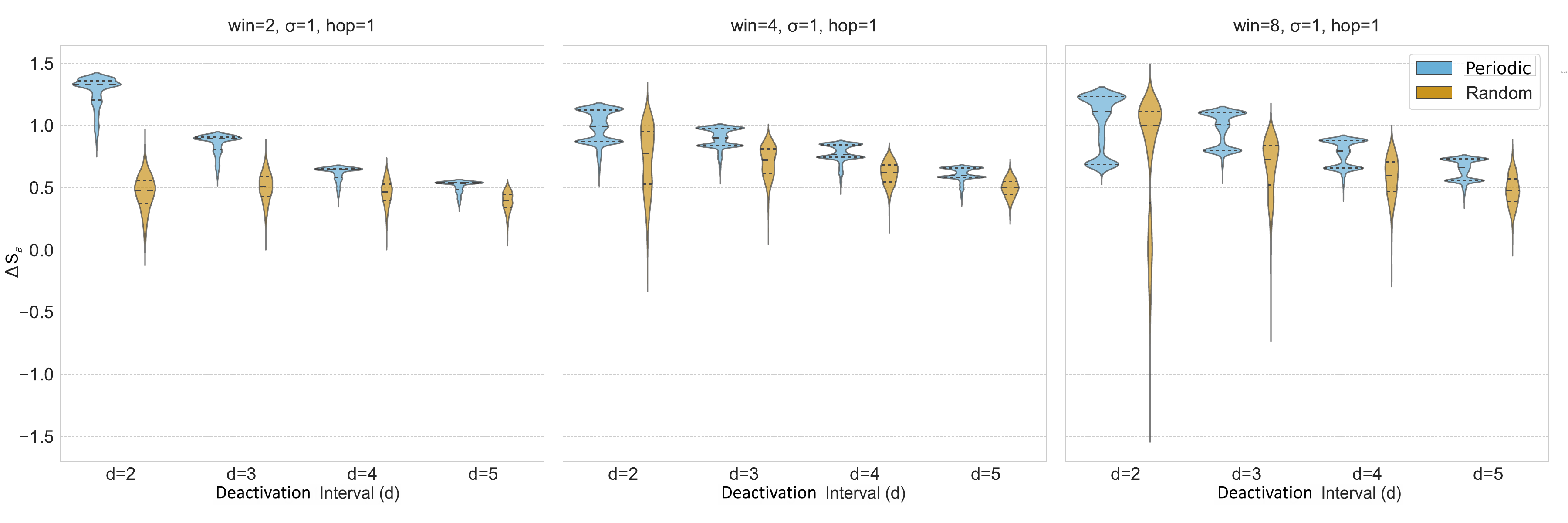}
    \caption{\textbf{Ablation of Husimi parameters on MIMO.}
    Each panel corresponds to a specific parameter configuration of win, $\sigma$, and hop. Across all configurations, the $\Delta S_\mathcal{B}$, exhibits a monotonic decrease as the deactivation interval $d$ increases. Furthermore, the \textsf{Periodic} deactivation scheme consistently yields a greater entropy perturbation than the \textsf{Random} scheme.}
    \label{fig:mimo-husimi-ablation}
\end{figure}

\subsection{MRI: Ablation on Husimi Parameters}

\begin{figure}[H]
    \centering
    \includegraphics[width=\textwidth]{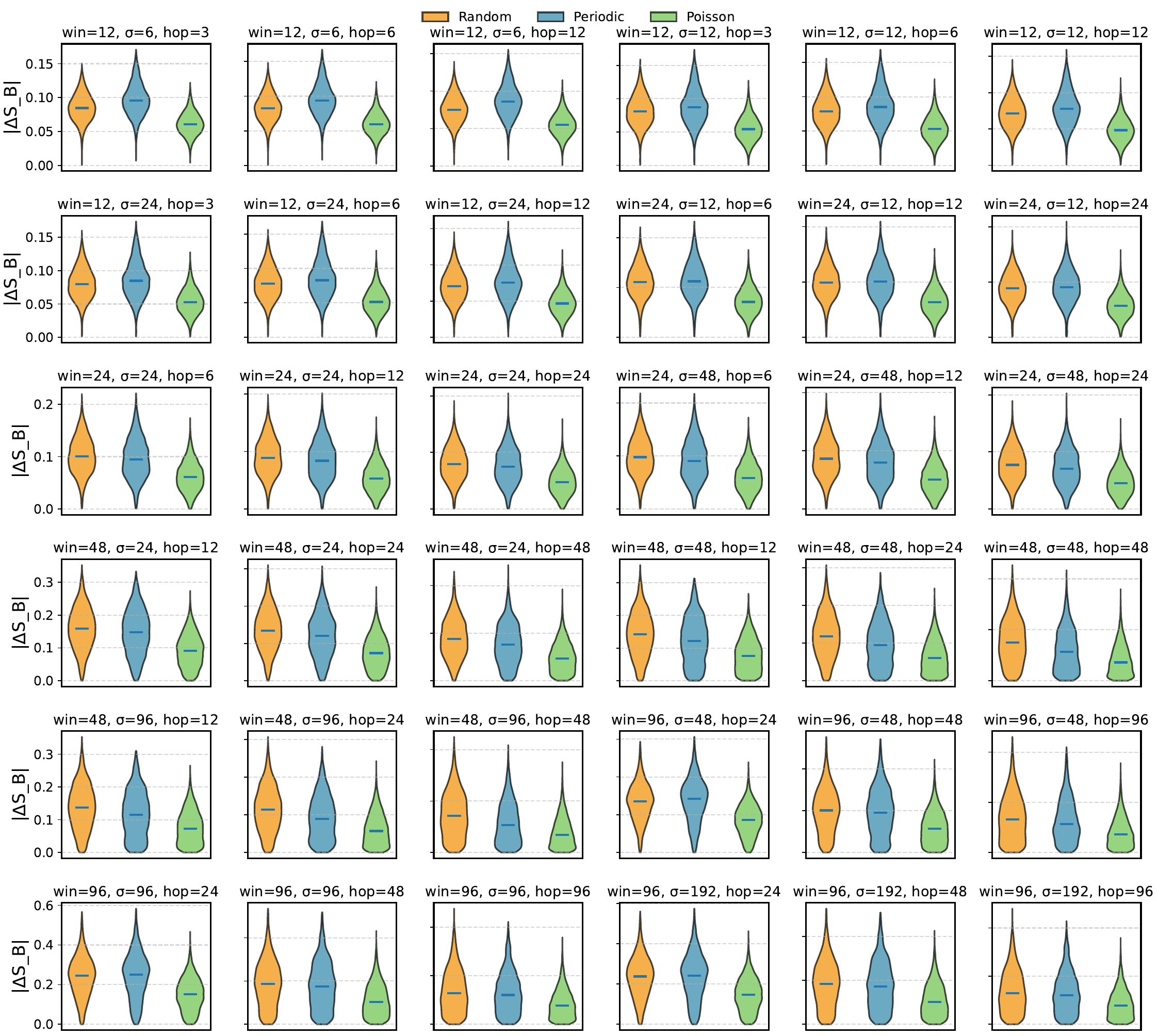}
    \caption{\textbf{Ablation of Husimi parameters on MRI.}
    Each subplot corresponds to one combination of $(\texttt{win},\sigma,\texttt{hop})$; violin plots summarize the distribution of $|\Delta S_{\mathcal B}|$ across all slices and acceleration ratios for three mask types (\textsf{Random}, \textsf{Periodic}, \textsf{Poisson}). 
    The overall ranking \textsf{Poisson}$<$\textsf{Periodic}$<$\textsf{Random} remains stable across nearly all parameter settings, consistent with the empirical reconstruction results in the main text. 
    The ordering partially reverses only at extremely small windows (\texttt{win}=12), where the Husimi window covers insufficient spatial–frequency support to distinguish the structured aliasing of \textsf{Periodic} from the stochastic loss in \textsf{Random}. 
    This highlights the importance of selecting Husimi parameters that are physically and architecturally consistent with both the imaging resolution and the receptive-field scale of the downstream network.}
    \label{fig:mri-husimi-ablation}
\end{figure}

To evaluate the stability of the Husimi-based entropy auditing in magnetic resonance imaging (MRI), we conducted an extensive grid search across the parameter space of the Gaussian window (\texttt{win}), its standard deviation ($\sigma$), and hop size (\texttt{hop}). 
For each parameter combination, the band-entropy change $|\Delta S_{\mathcal B}|$ was computed from the same set of fully sampled $k$-space data under three canonical mask types—\textsf{Random}, \textsf{Periodic}, and \textsf{Poisson}—and multiple acceleration ratios. 
Each configuration was processed using the same pipeline described in Main; the analysis was performed entirely prior to training, and the results were visualized as a $6\times6$ grid of violin plots (Fig.~\ref{fig:mri-husimi-ablation}).

We systematically varied \texttt{win}$\in\{12,24,48,96\}$, $\sigma/\texttt{win}\in\{0.5,1,2\}$, and $\texttt{hop}/\texttt{win}\in\{0.25,0.5,1\}$, producing $4\times3\times3=36$ configurations. 
For each combination, $|\Delta S_{\mathcal B}|$ values across all slices and sampling ratios were aggregated per mask type to produce violin plots that summarize their statistical distributions. 
All subplots share a common color code (orange = \textsf{Random}, blue = \textsf{Periodic}, green = \textsf{Poisson}) and display the parameter triple $(\texttt{win},\sigma,\texttt{hop})$ in their title; each panel’s vertical scale is locally adaptive to emphasize within-condition trends.

Across the majority of settings, the ordering of mean $|\Delta S_{\mathcal B}|$ values is highly stable:
\[
\textsf{Poisson} \;<\; \textsf{Periodic} \;<\; \textsf{Random}.
\]
This hierarchy exactly matches the empirical reconstruction ranking reported in the main text, demonstrating that the Husimi-based entropy measure faithfully captures the relative information preservation of different undersampling patterns. 
The \textsf{Poisson} mask consistently produces the smallest $|\Delta S_{\mathcal B}|$, indicating that its variable-density sampling preserves the original spectral organization most effectively.

When the window size is very small (\texttt{win}=12), however, the ordering between \textsf{Periodic} and \textsf{Random} reverses. 
In this regime, the Husimi window spans too few pixels to include sufficient structural information, and the entropy estimate becomes dominated by local intensity variations rather than global aliasing structure. 
Physically, this corresponds to an under-resolved phase-space measurement: the local spectrogram fails to capture the coherent folding patterns induced by periodic subsampling. 
From a representational perspective, the network’s convolutional receptive fields are relatively broad compared with such a small Husimi window, so the entropy measure loses sensitivity to the periodicity that the network can still exploit. 

These findings emphasize that the choice of Husimi parameters should reflect not only the imaging physics (spatial–frequency resolution of the acquisition) but also the representational scale of the downstream model. 
In other words, the Husimi transform serves as a conceptual bridge between the \emph{measurement domain} and the \emph{learning domain}: 
its parameters determine at what scale the instrument-resolved structure is compared to the model’s feature extraction capability. 
When chosen appropriately, the band-entropy change $|\Delta S_{\mathcal B}|$ provides a stable, physically interpretable predictor of reconstruction difficulty across undersampling strategies.

\section{Supporting Note 4: Quality–entropy correlation plots for MRI}
\label{sec:si-mri-quality-entropy}

To complement the main-text results, we explicitly
visualize the association between band-entropy change and reconstruction
quality for MRI. For each acquisition configuration (mask family and
acceleration factor), we compute:
\begin{itemize}
    \item the mean band-entropy change $|\Delta S_{\mathcal B}|$, averaged over all test slices, and
    \item the corresponding mean reconstruction metrics (PSNR and SSIM), again averaged over the same test slices.
\end{itemize}
This yields one point per configuration in the quality–entropy plane, with
each point summarizing the population behavior of a given mask at a given
acceleration factor.

\begin{figure*}[t]
    \centering
    \includegraphics[width=\textwidth]{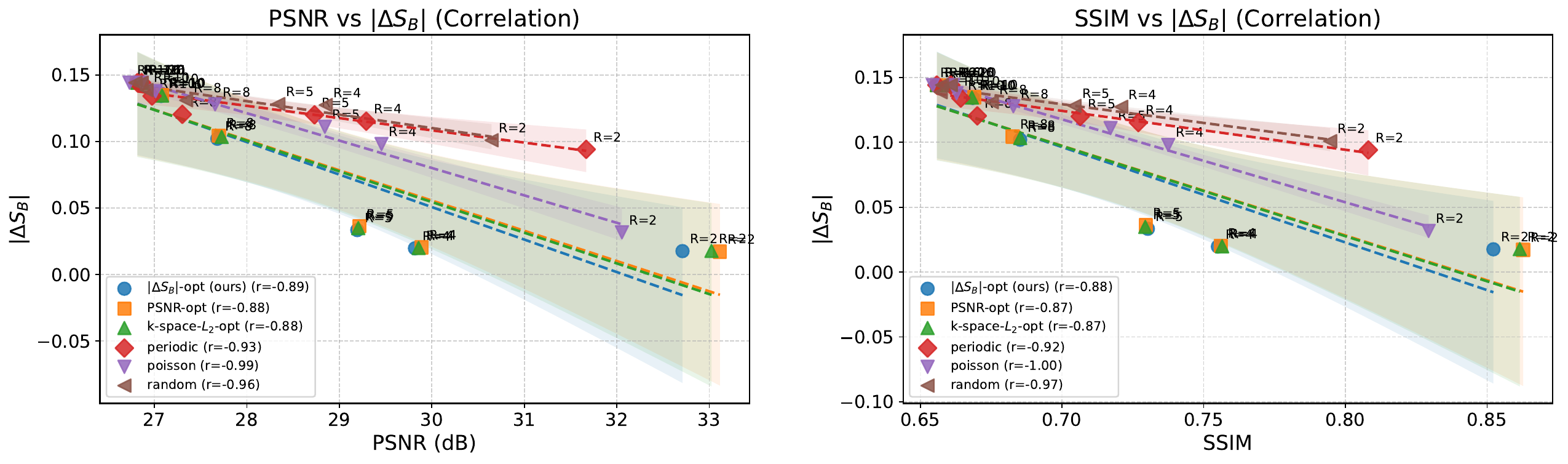}
    \caption{\textbf{Quality–entropy association for accelerated MRI.}
    Each point corresponds to one mask family at one acceleration factor,
    summarized by its mean band-entropy change $|\Delta S_{\mathcal B}|$ (horizontal axis)
    and mean reconstruction quality (vertical axis) over all test slices.
    Different colors denote mask families (Periodic, Random, Poisson, and
    three variable-density designs), and different marker shapes/alphas
    indicate acceleration factors.
    \textbf{(a)} Mean PSNR versus mean $|\Delta S_{\mathcal B}|$.
    \textbf{(b)} Mean SSIM versus mean $|\Delta S_{\mathcal B}|$.
    Dashed lines show ordinary least squares (OLS) fits across all
    configurations; shaded regions indicate $95\%$ confidence intervals for
    the mean regression line. Pearson r values are reported with 95\% confidence intervals estimated by Fisher’s z-transform. In both panels, configurations with smaller
    $|\Delta S_{\mathcal B}|$ tend to exhibit higher PSNR/SSIM, indicating that
    acquisition schemes that minimally perturb band entropy also tend to
    yield easier downstream reconstruction tasks.}
    \label{fig:correlation_analysis}
\end{figure*}

Figure~\ref{fig:correlation_analysis} shows scatter plots of mean
reconstruction quality versus mean $|\Delta S_{\mathcal B}|$ for all six mask
families (Periodic, Random, Poisson, and the three parametric
variable-density designs), across all acceleration factors considered in the
main text. Different colors encode mask families and different marker
shapes encode acceleration factors. Panel~(a) plots mean PSNR against mean
$|\Delta S_{\mathcal B}|$, while panel~(b) plots mean SSIM against the same entropy
change. In both panels, we fit an ordinary least squares (OLS) regression
line across all configurations (dashed line) and show the $95\%$ confidence
interval for the mean regression line (shaded band).

Despite the fact that PSNR and SSIM are already near saturation for several
variable-density designs, a clear monotone association emerges: configurations with smaller
$|\Delta S_{\mathcal B}|$ tend to achieve higher PSNR and SSIM. This trend is
consistent across all six mask families and all acceleration factors, and
no single family dominates the correlation.

These quality–entropy correlation plots therefore support the main-text
interpretation: in accelerated MRI, the band-entropy change $|\Delta S_{\mathcal B}|$,
computed \emph{before} any network is trained, is positively associated
with downstream reconstruction difficulty. Sampling designs that better
preserve the original Husimi band-entropy (smaller $|\Delta S_{\mathcal B}|$)
tend to yield easier reconstruction tasks as measured by PSNR and SSIM.

